\documentclass[runningheads]{llncs}
\usepackage[T1]{fontenc}
\usepackage{graphicx}
\usepackage{booktabs}
\usepackage[misc]{ifsym}
\newcommand{\corr}{(\Letter)}
\usepackage{mwe}
\usepackage{amsmath,amsfonts,bm}

\def\eqref#1{equation~\ref{#1}}
\def\1{\bm{1}}

\def\rvz{{\mathbf{z}}}

\def\vx{{\bm{x}}}
\def\vy{{\bm{y}}}

\def\mX{{\bm{X}}}

\DeclareMathAlphabet{\mathsfit}{\encodingdefault}{\sfdefault}{m}{sl}
\SetMathAlphabet{\mathsfit}{bold}{\encodingdefault}{\sfdefault}{bx}{n}

\def\sL{{\mathbb{L}}}

\def\sV{{\mathbb{V}}}

\usepackage{hyperref}
\usepackage{url}
\usepackage{graphicx}
\usepackage{booktabs} 
\usepackage{multirow}
\usepackage[capitalize,noabbrev]{cleveref}
\usepackage{comment}
\usepackage{todonotes}
\usepackage{mathtools}
\usepackage{wrapfig}

\newcommand{\cmark}{\checkmark}
\newcommand{\xmark}{}

\begin{document}

\title{TreeDiffusion: Hierarchical Generative Clustering for Conditional Diffusion}

\titlerunning{Hierarchical Clustering for Conditional Diffusion}
% If the full title of your paper is short enough to also fit in the running head, you can omit the abbreviated paper title here. You can check as follows: if you comment out the \titlerunning line, something will appear in the header of all odd-numbered pages of your PDF from page 3 onward. This something is either the full title (in which case all is well), or the error message "Title Suppressed Due to Excessive Length". If this error message appears, you're going to want to provide an abbreviated title within the \titlerunning command, because if you won't do it, Springer will do it for you.

%N.B.: Author information (both in the \author{} and \authorrunning{} command) should only be present in the Camera-Ready Version of your paper. The version that you initially submit for review, ought to be double-blind. So, when initially submitting your paper, use:
%\author{Author information scrubbed for double-blind reviewing}
\author{Jorge da Silva Gonçalves \corr \and
Laura Manduchi \and
Moritz Vandenhirtz \and \\
Julia E. Vogt} % \orcidID{0000-1111-2222-3333}}
% You may leave out the orcidID information, if you want to.
% Use \corr to indicate the corresponding author. Note the spacing around the \corr command. Only one author can be the corresponding author.

\tocauthor{Jorge da Silva Gonçalves, 
Laura Manduchi,
Moritz Vandenhirtz,
Julia E. Vogt}

\toctitle{TreeDiffusion: Hierarchical Generative Clustering for Conditional Diffusion}

%N.B.: comment out the \authorrunning{} command for the double-blind version of your paper submitted for review. Later, if your paper is accepted, use the command for the Camera-Ready Version.
\authorrunning{J. da Silva Gonçalves et al.}

% If there are two authors, write 'A.L. Benjamin and C.C. Broadus Jr.'
% If there are more than two authors, '[...] et al.' is used.

\institute{ETH Zurich, Rämistrasse 101, 8092 Zurich, Switzerland \email{jorge.dasilvagoncalves@inf.ethz.ch}}
% \email{\{jorge.dasilvagoncalves, laura.manduchi, moritz.vandenhirtz, julia.vogt\}@inf.ethz.ch} }

\maketitle              % typeset the header of the contribution

\begin{abstract}

Generative modeling and clustering are conventionally distinct tasks in machine learning. Variational Autoencoders (VAEs) have been widely explored for their ability to integrate both, providing a framework for generative clustering. However, while VAEs can learn meaningful cluster representations in latent space, they often struggle to generate high-quality samples. This paper addresses this problem by introducing TreeDiffusion, a deep generative model that conditions diffusion models on learned latent hierarchical cluster representations from a VAE to obtain high-quality, cluster-specific generations. Our approach consists of two steps: first, a VAE-based clustering model learns a hierarchical latent representation of the data. Second, a cluster-aware diffusion model generates realistic images conditioned on the learned hierarchical structure. We systematically compare the generative capabilities of our approach with those of alternative conditioning strategies. Empirically, we demonstrate that conditioning diffusion models on hierarchical cluster representations improves the generative performance on real-world datasets compared to other approaches. Moreover, a key strength of our method lies in its ability to generate images that are both representative and specific to each cluster, enabling more detailed visualization of the learned latent structure. Our approach addresses the generative limitations of VAE-based clustering approaches by leveraging their learned structure, thereby advancing the field of generative clustering.

\end{abstract}

% -----------------------------------------------------------------------------

This preprint has not undergone peer review or any post-submission improvements or corrections.
The Version of Record of this contribution is published in Machine Learning and Knowledge Discovery in Databases. Research Track. ECML PKDD 2025, Lecture Notes in Computer Science (LNCS, vol. 16013), and is available online at \url{https://doi.org/10.1007/978-3-032-05962-8_26}.

\newpage

\section{Introduction}

Generative modeling and clustering are two fundamental yet different tasks in machine learning. Generative modeling focuses on approximating the underlying data distribution, enabling the generation of new samples \cite{kingma_auto-encoding_2013,goodfellow_generative_2014}. Clustering, on the other hand, seeks to uncover meaningful and interpretable structures within data through the unsupervised detection of intrinsic relationships and dependencies \cite{ward1963hierarchical,ezugwu_comprehensive_2022}, facilitating better visualization and interpretation of the data.
By integrating hierarchical dependencies into a deep latent variable model, TreeVAE \cite{manduchi_tree_2023} was recently proposed to bridge these two research directions. While TreeVAE is effective at hierarchical clustering, it falls short in generating high-quality images. Like other VAE-based models, it faces common issues such as producing blurry outputs \cite{bredell_explicitly_2023}. In contrast, diffusion models \cite{sohl-dickstein_deep_2015,ho_denoising_2020} have recently gained prominence for their superior image generation capabilities, progressively refining noisy inputs to produce sharp, realistic images.

Our work bridges this gap by introducing a second-stage diffusion model that is conditioned on the hierarchical cluster representations learned by TreeVAE. The proposed framework, \textbf{TreeDiffusion}, combines the strengths of both models to generate high-quality, cluster-specific images, achieving improved performance in image generation. 
The generative process begins by sampling the root embedding of a latent tree, which is learned during training. 
From there, the sample is propagated from the root to one leaf by (a) sampling a path through the tree and (b) applying a sequence of stochastic transformations to the root embedding along the chosen hierarchical path.
Subsequently, the diffusion model harnesses the hierarchical information by conditioning its reverse diffusion process on the sampled path representations of the latent tree through a path encoder. A key strength of TreeDiffusion is its ability to generate images tailored to each cluster, providing enhanced visualization of the learned representations, as demonstrated by our qualitative results. For the same sample, our method can produce leaf-specific images that share common general properties but differ by features encoded in the latent hierarchy. Lastly, this approach overcomes the generative limitations of VAE-based hierarchical clustering models like TreeVAE while preserving their clustering performance.

\subsection{Main Contributions}

Our main contributions include (i) a unified framework that integrates hierarchical clustering into diffusion models, and (ii) a novel mechanism for controlling image synthesis based on learned clusters. We demonstrate that our approach (a) surpasses the generative limitations of VAE-based clustering models, and (b) produces samples that are both more representative of their respective clusters and closer to the true data distribution than models without hierarchical clustering integration.

% -----------------------------------------------------------------------------

%% Will need to change the wording in Related Work section because of the citation format, to avoid using [number] as nouns in sentence
 
\section{Related Work}
\subsubsection{Variational Approaches for Hierarchical Clustering.}
Since their introduction, Variational Autoencoders (VAEs) \cite{kingma_auto-encoding_2013} have been widely used for clustering tasks, due to their ability to learn structured latent representations \cite{jiang_variational_2017}. For instance, one method integrates hierarchical Bayesian non-parametric priors to the latent space of VAEs by applying the nested Chinese Restaurant Processes to cluster the data based on infinitely deep and branching trees \cite{goyal_nonparametric_2017}. Other methods construct hierarchical representations during training, such as DeepECT \cite{mautz_deepect_2020} and TreeVAE \cite{manduchi_tree_2023}. While DeepECT aggregates data into a hierarchical tree in a single shared latent space, TreeVAE learns a tree-structured posterior distribution of latent stochastic variables. That is, TreeVAE models the data distribution by learning an optimal tree structure of latent variables. This results in latent embeddings that are automatically organized into a hierarchy, mimicking the hierarchical clustering process. Single-cell TreeVAE \cite{vandenhirtz2024sctree} further extends TreeVAE to single-cell RNA sequencing data by integrating batch correction, facilitating biologically plausible hierarchical structures. Despite their strong clustering performance, these models often fall short in generative quality, with few providing quantitative or qualitative evaluations of their sample generation capabilities.

% Will most likely shorten the "Diffusion Models" paragraph and merge it with the "Connecting Diffusion with Clustering" paragraph

\subsubsection{Diffusion Models.}
Diffusion models have become state-of-the-art for image generation tasks over the past few years \cite{sohl-dickstein_deep_2015,ho_denoising_2020,song_denoising_2020,nichol_improved_2021,dhariwal_diffusion_2021,vahdat_score-based_2021,DBLP:conf/iclr/SalimansH22,rombach_high-resolution_2022}. One drawback of diffusion models is that their latent variables lack interpretability compared to the latent spaces of VAEs. To take advantage of the strengths of both approaches, researchers have explored architectures that combine the more interpretable latent spaces of VAEs with the advanced generative capabilities of diffusion models. Notable examples include DiffuseVAE \cite{pandey_diffusevae_2022}, Diffusion Autoencoders \cite{preechakul_diffusion_2022}, and InfoDiffusion \cite{wang_infodiffusion_2023}. Representation-conditioned image generation \cite{li2023self} illustrates how self-supervised learning can improve generative diffusion frameworks in unsupervised settings, reducing the gap between class-conditional and unconditional image generation.

\subsubsection{Connecting Diffusion with Clustering.}
The research most closely related to our work focuses on using clustering as conditioning signals for diffusion models to improve generative quality. One approach \cite{adaloglou2024rethinking} utilizes cluster assignments from k-means or TEMI clustering \cite{DBLP:conf/bmvc/AdaloglouMKK23}. Similarly, another one \cite{hu2023self} introduces a framework that employs the k-means clustering algorithm as an annotation function, generating self-annotated image-level, box-level, and pixel-level guidance signals. Both studies demonstrate the benefits of conditioning on clustering information to improve generative performance without going into the specifics of clustering performance itself. In contrast, our work further investigates which types of clustering information are most beneficial for the model, employing learned latent cluster representations alongside cluster assignments for conditioning. Related to conditioning on clusters, both kNN-Diffusion \cite{DBLP:conf/iclr/SheyninAPSGNT23} and Retrieval-Augmented Diffusion Models \cite{blattmann2022retrieval} utilize nearest neighbor retrieval to condition generative models on similar embeddings, minimizing the need for large parametric models and paired datasets in tasks like text-to-image synthesis. Diffusion models have also been applied in incomplete multiview clustering to generate missing views to improve clustering performance \cite{DBLP:conf/icml/0001DWC0F024,wen2020generalized}. %Recent works \cite{liu2023unsupervised,su2024compositional} analyze the capability of diffusion models for unsupervised concept discovery, wherein image datasets are decomposed into meaningful compositional representations, similar to clustering. 
On a different note, recent research shows that training diffusion models is equivalent to solving a subspace clustering problem, explaining their ability to learn image distributions with few samples \cite{wang2024diffusion}. Finally, diffusion models have also been applied as a post-hoc method to enhance the generation quality of their multimodal clustering models \cite{palumbo_deep_2023}. However, to the best of our knowledge, no existing diffusion model explicitly uses hierarchical clustering to enhance the interpretability and generative performance of generative clustering models.

% -----------------------------------------------------------------------------

\section{Method}
\label{model}

\begin{figure*} %[ht]
\begin{center}
\centerline{\includegraphics[width=\textwidth, trim=40 5 50 5, clip]{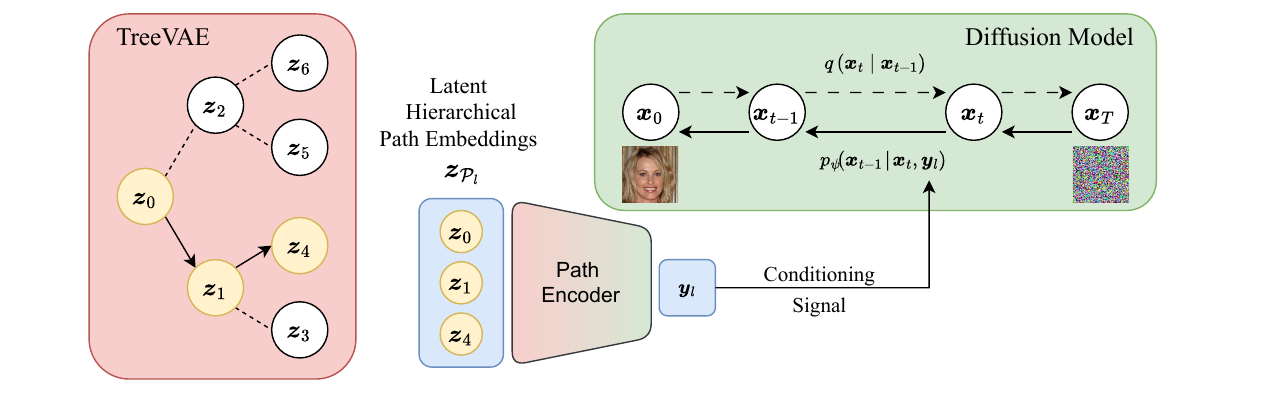}}
\caption[Schematic overview of the TreeDiffusion model]{Schematic overview of the TreeDiffusion framework: TreeVAE encodes data into hierarchical latent variables, where a path is sampled from the root to a leaf node. An encoder network creates a conditioning signal using the sampled hierarchical path embeddings. The diffusion model leverages this information to condition its reverse process and generate a cluster-specific image.}
\label{diff-treevae}
\end{center}
\end{figure*}

We propose \textbf{TreeDiffusion}
%\footnote{The code will be published upon acceptance.}, 
\footnote{The code is publicly available at \url{https://github.com/JoGo175/TreeDiffusion}.}, 
a two-stage framework consisting of a VAE-based generative hierarchical clustering model, followed by a hierarchy-conditional diffusion model. In the first stage, TreeVAE \cite{manduchi_tree_2023} serves as the clustering model, encoding hierarchical clusters within its latent tree structure, where the leaf nodes represent clusters. We select TreeVAE as it provides structured hierarchical latent representations from root to leaf, which are then processed by a path encoder to create the conditioning signal. In the second stage, a denoising diffusion implicit model (DDIM) \cite{pandey_diffusevae_2022}, uses this conditioning signal to generate cluster-conditional samples. Hence, our model enables cluster-guided diffusion in unsupervised settings, analogously to classifier-guided diffusion \cite{dhariwal_diffusion_2021} in supervised settings. \cref{diff-treevae} illustrates the workflow of TreeDiffusion. The following sections provide a detailed description of each stage of the model.

% need to better explain, that it is only the latent structure of the TreeVAE in image

\subsection{Hierarchical Clustering with TreeVAE}

The first part of TreeDiffusion involves a Tree Variational Autoencoder (TreeVAE) \cite{manduchi_tree_2023}. TreeVAE is a generative model that learns to hierarchically separate data into clusters through a latent tree structure. During training, the model dynamically grows a binary tree structure of stochastic variables. The process begins with a simple tree composed of a root and two child nodes, and it optimizes the corresponding ELBO over a fixed number of epochs. Afterward, the tree expands by adding two child nodes to an existing leaf node, prioritizing nodes with the highest assigned sample count to promote balanced leaves. This expansion continues iteratively, training only the subtree formed by the new leaves while freezing the rest of the model. This process repeats until the tree reaches a predefined depth or leaf count, alternating between optimizing model parameters and expanding the tree structure.

Let the set $\sV$ represent the nodes of the tree. Each node corresponds to a stochastic latent variable, denoted as $\rvz_0, \dots, \rvz_V$. Each latent variable follows a Gaussian distribution, whose parameters depend on their parent nodes through neural network layers. The set of leaves $\sL$, with $\sL \subset \sV$, represents the clusters present in the data. Starting from the root node, $\rvz_0$, a given sample traverses the tree to a leaf node, $\rvz_l$, in a probabilistic manner. The probabilities of moving to the left or right child at each internal node are determined by neural networks termed \textit{routers}. These decisions, denoted by $\mathrm{c}_i$ for each non-leaf node $i$, follow a Bernoulli distribution, where $\mathrm{c}_i = 0$ indicates the selection of the left child. The path $\mathcal{P}_l$ refers to the sequence of nodes from the root to one leaf $l$. Moreover, let $\boldsymbol{z}_{\mathcal{P}_l}=\left\{\boldsymbol{z}_i \mid i \in \mathcal{P}_l\right\}$ denote the set of latent embeddings for each node in the path $\mathcal{P}_l$. The latent tree encodes a sample-specific probability distribution of paths. Each leaf embedding, $\rvz_l$ for $l \in \sL$, represents the learned data representations. In TreeVAE, leaf-specific decoders use these embeddings to reconstruct or generate new cluster-specific images, i.e. given a dataset $\mX$, TreeVAE reconstructs $\hat{\mX} = \{\hat{\mX}^{(l)} \mid l \in \sL\}$. The generative model (\ref{eq:TreeVAE_gen_model}) and inference model (\ref{eq:TreeVAE_inf_model}) of TreeVAE are defined as follows, according to \cite{manduchi_tree_2023}:
\begin{align}
    p_\theta\left(\boldsymbol{z}_{\mathcal{P}_l}, \mathcal{P}_l\right) &= p\left(\boldsymbol{z}_0\right) \prod_{i \in \mathcal{P}_l \backslash\{0\}} \underbrace{p\left(\mathrm{c}_{p a(i) \rightarrow i} \mid \boldsymbol{z}_{p a(i)}\right)}_{\text {decision probability}} \underbrace{p\left(\boldsymbol{z}_i \mid \boldsymbol{z}_{p a(i)}\right)}_{\text {sample probability}} \label{eq:TreeVAE_gen_model} \\
    q\left(\boldsymbol{z}_{\mathcal{P}_l}, \mathcal{P}_l \mid \boldsymbol{x}\right) &= q\left(\boldsymbol{z}_0 \mid \boldsymbol{x}\right) \prod_{i \in \mathcal{P}_l \backslash\{0\}} q\left(\mathrm{c}_{p a(i) \rightarrow i} \mid \boldsymbol{x}\right) q\left(\boldsymbol{z}_i \mid \boldsymbol{z}_{p a(i)}\right) \label{eq:TreeVAE_inf_model}
\end{align}

\subsection{Diffusion conditioned on Hierarchical Clusters} % Change title, improve it

The second part of TreeDiffusion incorporates a conditional diffusion model. We assume the same forward process as in standard Denoising Diffusion Probabilistic Models (DDPM) \cite{ho_denoising_2020}, which gradually introduces noise to the data $\boldsymbol{x}_0$ over $T$ steps. The intermediate states, $\boldsymbol{x}_t$ for $t = 1, \dots, T$, follow a trajectory determined by a noise schedule ${\beta_1, \dots, \beta_T}$ that controls the rate of data degradation, whereby $\alpha_t=\left(1-\beta_t\right)$ and $\bar{\alpha}_t=\prod_{s=1}^t \alpha_s$. Hence, the forward process can be summarized as follows:
\begin{align}
q\left(\boldsymbol{x}_{1: T} \mid \boldsymbol{x}_0\right)&=\prod_{t=1}^T q\left(\boldsymbol{x}_t \mid \boldsymbol{x}_{t-1}\right) \\
q\left(\boldsymbol{x}_t \mid \boldsymbol{x}_{t-1}\right)&=\mathcal{N}\left(\sqrt{1-\beta_t} \boldsymbol{x}_{t-1}, \beta_t \boldsymbol{I}\right) \\ 
q\left(\boldsymbol{x}_t \mid \boldsymbol{x}_0\right)&=\mathcal{N}\left(\sqrt{\bar{\alpha}_t} \boldsymbol{x}_0,\left(1-\overline{\alpha}_t\right) \boldsymbol{I}\right)
\end{align}

For the reverse process, TreeDiffusion starts with random noise, similar to standard diffusion models. TreeDiffusion relies exclusively on the latent hierarchical information provided by TreeVAE, which is based on the tree structure learned during the first stage. Specifically, a tree leaf $l$ corresponds to a selected cluster, determined by sampling from the TreeVAE path probabilities. Thus, this leaf represents a unique path through the hierarchical structure. The hierarchical conditioning information is derived from $\boldsymbol{z}_{\mathcal{P}_l}$, the set of latent embeddings along the path from the root node to the chosen leaf. These embeddings are further processed by a dedicated path encoder, which aggregates the information to produce the conditioning signal $\vy_l$:
$$
\vy_l =
\sum_{i \in \mathcal{P}_l} \left( f_{\text{embed}}(\boldsymbol{z}_i) + f_{\text{node}}(i) \right).
$$

Here, $f_{\text{embed}}$ and $f_{\text{node}}$ are each implemented as projection blocks consisting of two MLP layers with an activation in-between, and jointly trained with the diffusion model. For each node in the path, its embedding and corresponding node index are projected independently into the time embedding dimension of the U-Net decoder \cite{ronneberger_u-net_2015,nichol_improved_2021}. The resulting projections are aggregated into the unified conditioning signal $\vy_l$, which is then combined with the time-step embeddings to guide the U-Net during the denoising process. Consequently, this conditioning mechanism directly influences the reverse process. Let $\psi$ denote the parameters of the denoising model, and let $p(l | \vx_0)$ be the probability that the sample $\vx_0$ is assigned to leaf $l$ in the latent tree. The reverse process can then be summarized as follows:
\begin{align}
    \begin{split} 
    &l \sim p(l | \vx_0), \\
    &p_{\psi} \!\! \left(\vx_{0: T} \!\mid \!\vy_l \right)=p\left(\vx_{T}\right) \prod_{t=1}^T p_\psi \!\!\left(\vx_{t-1} \!\mid \!\vx_{t}, \vy_l \right)\!, \\ 
% \begin{split} &p_\psi\left(\vx_{t-1} \mid \vx_{t}, \hat{\vx}_{0}^{(l)}, l \right)= \\
% &\mathcal{N}\left(\mu_\psi\left(\vx_{t}, t, \hat{\vx}_{0}^{(l)}, l\right), \Sigma_\psi\left(\vx_{t}, t, \hat{\vx}_{0}^{(l)}, l\right)\right), 
% & \;\;\;\;\; \text{ with } l \sim p(l | \vx_0)
\end{split}
\end{align}

The path sampling ensures that different leaves are considered, encouraging the diffusion model to perform effectively across all leaves. Consequently, our approach addresses the distinct clusters inherent to TreeVAE, allowing the model to adapt and encouraging cluster-aware refinements in the images. This guidance in the image generation process assists the denoising model in learning cluster-specific image reconstructions. 

The following design considerations are implemented in TreeDiffusion to achieve computational efficiency without compromising effectiveness. Due to the large number of denoising steps required, DDPM sampling can be computationally expensive. To address this issue, we opt for the DDIM sampling procedure \cite{song_denoising_2020} instead of the standard DDPM \cite{ho_denoising_2020}. DDIMs significantly accelerate inference by using only a subset of denoising steps, making the process more efficient while maintaining high-quality results. Finally, by employing a two-stage training strategy, where the conditional diffusion model is trained using a pre-trained TreeVAE model, TreeDiffusion preserves the hierarchical clustering performance of TreeVAE. Hence, we can combine the effective clustering of TreeVAE with the superior image generation capabilities of diffusion models.

% -----------------------------------------------------------------------------

%\section{Experiments}
%\subsection{Experimental Setup}
%\subsection{Experimental Results}

% TODO: adapt Experiments section, think about what goes into the discussion

\section{Experiments}
\label{experiments}

We present a series of experiments to evaluate the performance of TreeDiffusion across various datasets. The experiments are carried out on MNIST \cite{lecun_gradient-based_1998}, FashionMNIST \cite{xiao_fashion-mnist_2017}, CIFAR-10 \cite{krizhevsky_learning_2009}, CelebA \cite{liu2015faceattributes}, and CUBICC \cite{palumbo_deep_2023}. The CUBICC dataset is a variant of the CUB Image-Captions dataset \cite{wah_branson_welinder_perona_belongie_2011,shi_variational_2019}, consisting of bird images grouped into eight distinct species. In Section \ref{performance}, we compare the generative performance of TreeDiffusion against baseline methods. In Section \ref{cluster_specifivity}, we evaluate how specific the generated images are to their source clusters and analyze how distinct images from different clusters are from one another. Finally, in Section \ref{ablation}, we perform an ablation study on conditioning signals, examining various model configurations to identify which signals most effectively improve generative performance. Specifically, we compare conditioning strategies based on hierarchical clustering, flat clustering, and no clustering information.

\subsection{Generative Performance} \label{performance}

The following analysis compares the proposed TreeDiffusion model with TreeVAE \cite{manduchi_tree_2023}, and a naive hybrid approach — referred to as "TreeVAE + Diffusion". In this hybrid approach, the diffusion model refines the TreeVAE output and refines it but without being conditioned on any latent information, as it is done in DiffuseVAE \cite{pandey_diffusevae_2022}. In contrast, TreeDiffusion conditions the diffusion model on the path information of the selected cluster retrieved from TreeVAE. For both these models, the TreeVAE instances serve as the first-stage model. % To ensure a fair comparison, we maintain comparable parameter counts for the respective VAE components across all compared methods, as well as for the diffusion components where applicable. 
The evaluation considers both reconstruction and generative performance, measured using the Fréchet Inception Distance (FID) \cite{heusel_gans_2017}. Reconstruction performance is assessed by computing the FID score between reconstructed images and their corresponding test set images. Generative performance is evaluated by calculating the FID score for 10,000 newly generated images. The results of this analysis are summarized in Table \ref{tab:rec_gen_performance}.

\begin{figure*}[h!]
\begin{center}
\centerline{\includegraphics[width=\textwidth]{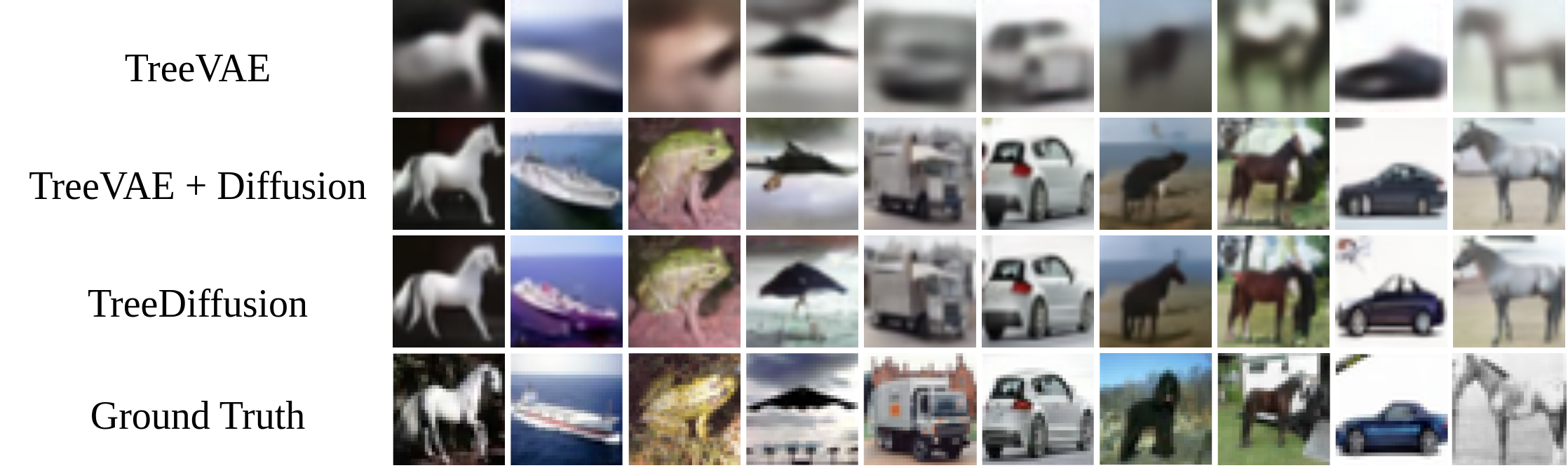}}
\caption{Ten different CIFAR-10 reconstructions generated by the TreeVAE model, each obtained by sampling a single path in the tree. Corresponding reconstructions from TreeVAE + Diffusion, which begins denoising with the TreeVAE reconstructions, are shown alongside those from TreeDiffusion, which conditions on the same selected path and embeddings but starts denoising from noise.}
\label{true_recons_comp}
\end{center}
\end{figure*}

The naive approach (TreeVAE + Diffusion) and TreeDiffusion both achieve substantial improvements over the baseline TreeVAE, their first-stage model, reducing FID scores by approximately an order of magnitude across all datasets. The naive approach performs better on simpler grayscale datasets, primarily excelling at image reconstruction rather than generation. This highlights a tendency toward overfitting. Conversely, TreeDiffusion consistently outperforms on the more complex, real-world color datasets at generating new images. Most likely, the difference in performance stems from how the denoising process is initialized. The naive model begins denoising from TreeVAE reconstructions, thereby making it highly dependent on the reconstruction quality provided by TreeVAE. Given that TreeVAE struggles more with generating new images than with reconstruction, this limitation is propagated into the naive approach. TreeDiffusion circumvents this issue by initializing the denoising directly from noise, using only latent representations from TreeVAE. As a result, TreeDiffusion achieves a better balance between reconstruction and generation quality, leading to better FID scores on newly generated images. \Cref{true_recons_comp} compares image reconstructions on CIFAR-10, demonstrating that both diffusion-based models significantly improve upon the image quality produced by TreeVAE.

% Compared to the TreeVAE-based approaches, DiffuseVAE underperforms on grayscale datasets but achieves better results on CIFAR-10 and CUBICC. However, TreeDiffusion outperforms DiffuseVAE in terms of generative FID on CelebA. The primary difference between these models lies in their VAE structures. TreeVAE struggles with generating high-quality images, which suggests that optimization might be easier when the model starts with better initial reconstructions from a standard VAE. Further investigation is required to explore this in future research.

% TODO
% ADD AN IMAGE COMPARISON HERE FOR SAMPLES

% This improvement is particularly evident in the quality of the generated images, as illustrated in \cref{recons_comp} for the CUBICC dataset. Here, we visually compare the samples generated by TreeDiffusion with those produced by the underlying TreeVAE, which supplied the cluster path as the conditioning signal for TreeDiffusion. The TreeVAE model continues to produce visibly blurry images, whereas TreeDiffusion generates noticeably sharper samples that adhere better to the data distribution. %Moreover, the TreeDiffusion model excels in reproducing fine details while maintaining the overall color and structure of the images. 

\begin{table}[h!]
\centering
\caption[Test set generative performances]{Test set generative performances of the compared models. FID scores for $10,000$ samples (lower is better) computed across $10$ random model initializations.}
\label{tab:rec_gen_performance}

    \begin{tabular*}{\linewidth}{@{\extracolsep{\fill}}llcc}
    \toprule
    \textbf{Dataset} & \textbf{Method} & \textbf{FID (rec)} $\downarrow$ & \textbf{FID (gen)} $\downarrow$ \\
    \hline \multirow{2}{*}{MNIST}    
    & TreeVAE                   & $ 24.0 \,\text{{\footnotesize $\pm\, 0.9$}}$                      
                                & $ 21.8 \,\text{{\footnotesize $\pm\, 0.7$}}$ \\
    % & DiffuseVAE                & $ 1.5 \,\text{{\footnotesize $\pm\, 0.0$}}$                       
    %                             & $ 18.9 \,\text{{\footnotesize $\pm\, 7.1$}}$ \\
    & TreeVAE + Diffusion       & $ \textbf{1.4} \,\text{{\footnotesize $\pm\, 0.0$}}$              
                                & $ \textbf{1.8} \,\text{{\footnotesize $\pm\, 0.1$}}$ \\
    & TreeDiffusion             & $ 1.5 \,\text{{\footnotesize $\pm\, 0.0$}}$                       
                                & $ \textbf{1.8} \,\text{{\footnotesize $\pm\, 0.1$}}$ \\
    \hline \multirow{2}{*}{Fashion}     
    & TreeVAE                   & $ 40.7 \,\text{{\footnotesize $\pm\, 2.1$}}$                      
                                & $ 41.9 \,\text{{\footnotesize $\pm\, 2.1$}}$ \\
    % & DiffuseVAE                & $ \textbf{4.5} \,\text{{\footnotesize $\pm\, 0.2$}}$              
    %                             & $ 12.0 \,\text{{\footnotesize $\pm\, 4.9$}}$ \\
    & TreeVAE + Diffusion       & $ \textbf{4.8} \,\text{{\footnotesize $\pm\, 0.2$}}$                       
                                & $ \textbf{4.8} \,\text{{\footnotesize $\pm\, 0.2$}}$ \\
    & TreeDiffusion             & $ 5.5 \,\text{{\footnotesize $\pm\, 0.6$}}$                       
                                & $ 5.4 \,\text{{\footnotesize $\pm\, 0.4$}}$ \\
    \hline \multirow{2}{*}{CIFAR-10}  
    & TreeVAE                   & $ 175.8 \,\text{{\footnotesize $\pm\, 1.4$}}$                     
                                & $ 188.0 \,\text{{\footnotesize $\pm\, 2.0$}}$ \\
    % & DiffuseVAE                & $ 12.8 \,\text{{\footnotesize $\pm\, 0.2$}}$                      
    %                             & $ \textbf{13.7} \,\text{{\footnotesize $\pm\, 0.4$}}$ \\
    & TreeVAE + Diffusion       & $ \textbf{12.3} \,\text{{\footnotesize $\pm\, 0.1$}}$                      
                                & $ 19.7 \,\text{{\footnotesize $\pm\, 0.2$}}$ \\
    & TreeDiffusion             & $ 12.5 \,\text{{\footnotesize $\pm\, 0.4$}}$                      
                                & $ \textbf{17.8} \,\text{{\footnotesize $\pm\, 0.4$}}$  \\
    \hline \multirow{2}{*}{CUBICC}  
    & TreeVAE                   & $ 232.5 \,\text{{\footnotesize $\pm\, 7.1$}}$                     
                                & $ 255.3 \,\text{{\footnotesize $\pm\, 8.8$}}$ \\
    % & DiffuseVAE                & $ 18.7 \,\text{{\footnotesize $\pm\, 2.2$}}$       
    %                             & $ \textbf{24.2} \,\text{{\footnotesize $\pm\, 4.1$}}$ \\
    & TreeVAE + Diffusion       & $ \textbf{12.7} \,\text{{\footnotesize $\pm\, 6.5$}}$                              
                                & $ 96.0 \,\text{{\footnotesize $\pm\, 2.1$}}$ \\
    & TreeDiffusion             & $ 13.4 \,\text{{\footnotesize $\pm\, 0.9$}}$                      
                                & $ \textbf{29.0} \,\text{{\footnotesize $\pm\, 5.4$}}$  \\
    \hline \multirow{2}{*}{CelebA}  
    & TreeVAE                   & $ 75.2 \,\text{{\footnotesize $\pm\, 15.0$}}$                     
                                & $ 77.9 \,\text{{\footnotesize $\pm\, 5.6$}}$ \\
    % & DiffuseVAE                & $ 14.9 \,\text{{\footnotesize $\pm\, 5.1$}}$   
    %                             & $ 24.1 \,\text{{\footnotesize $\pm\, 9.2$}}$ \\
    & TreeVAE + Diffusion       & $ 15.4 \,\text{{\footnotesize $\pm\, 3.2$}}$                              
                                & $ 30.1 \,\text{{\footnotesize $\pm\, 7.5$}}$ \\
    & TreeDiffusion             & $ \textbf{14.1} \,\text{{\footnotesize $\pm\, 6.0$}}$                      
                                & $ \textbf{18.4} \,\text{{\footnotesize $\pm\, 7.2$}}$ \\
    \bottomrule
    \end{tabular*}
\end{table}

\begin{figure*}[ht]
\begin{center}
\centerline{\includegraphics[width=\textwidth]{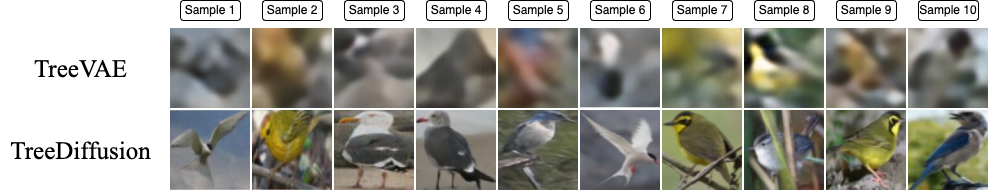}}
\caption{Ten different samples generated by the TreeVAE model, each generated by sampling one path in the tree, and corresponding samples from the TreeDiffusion model, conditioned on the same selected path and embeddings from TreeVAE.}
\label{recons_comp}
\end{center}
\end{figure*}
\begin{figure*}[h!]
\begin{center}
\centerline{\includegraphics[width=\textwidth, trim=0 30 20 0, clip]{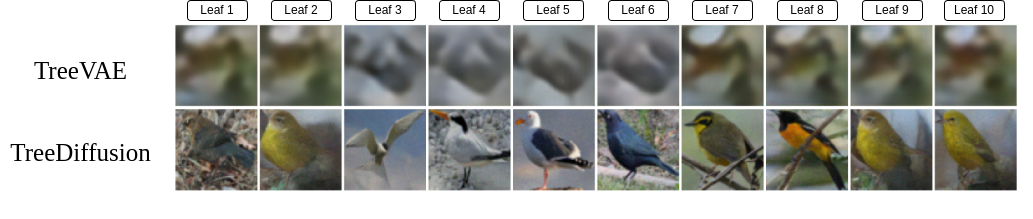}}
\caption{Image generations from every leaf of the TreeVAE and TreeDiffusion model, both trained on the CUBICC dataset. Each row shows the generated images from all leaves of the respective model, starting with the same root sample. }
\label{leaf_samples}
\end{center}
\end{figure*}

\subsection{Cluster-specific Representations} \label{cluster_specifivity}

\subsubsection{Higher quality cluster-specific generations.} In \cref{recons_comp}, we present randomly generated images for the CUBICC dataset for both TreeVAE and TreeDiffusion, where each column corresponds to an independently generated sample. For each generation, we first sample the root embedding; then, we sample the path in the tree and the refined representations along the selected path iteratively until a leaf is reached. The hierarchical representation is then used to condition the inference in TreeDiffusion. As can be seen, the TreeDiffusion generations show substantially higher generative quality. In the following, we examine the first generated sample from \cref{recons_comp} in more detail. For this one sample, we present the generations of all leaves in \cref{leaf_samples} by propagating the corresponding root representation across all paths in the tree. Note that the selected sample shown in \cref{recons_comp} ended up stemming from leaf 3 in \cref{leaf_samples}. When comparing the generated images across the leaves for both models, it is evident that TreeDiffusion not only produces sharper images for all clusters but also generates a greater diversity of images. Note that both models utilize the same latent information for image generation. While TreeVAE and TreeDiffusion maintain similar overall color distribution and structural characteristics, TreeDiffusion significantly enhances cluster specificity, resulting in images with greater clarity and distinctiveness for each cluster.

%Further examples of the leaf-specific image generations for TreeVAE and TreeDiffusion can be found in \ref{further_gen_examples}.

% need to find out how to link appendix, need to send it separately

\begin{figure*}[ht!]
\begin{center}
\centerline{\includegraphics[width=0.9\textwidth, trim=150 30 150 150, clip]{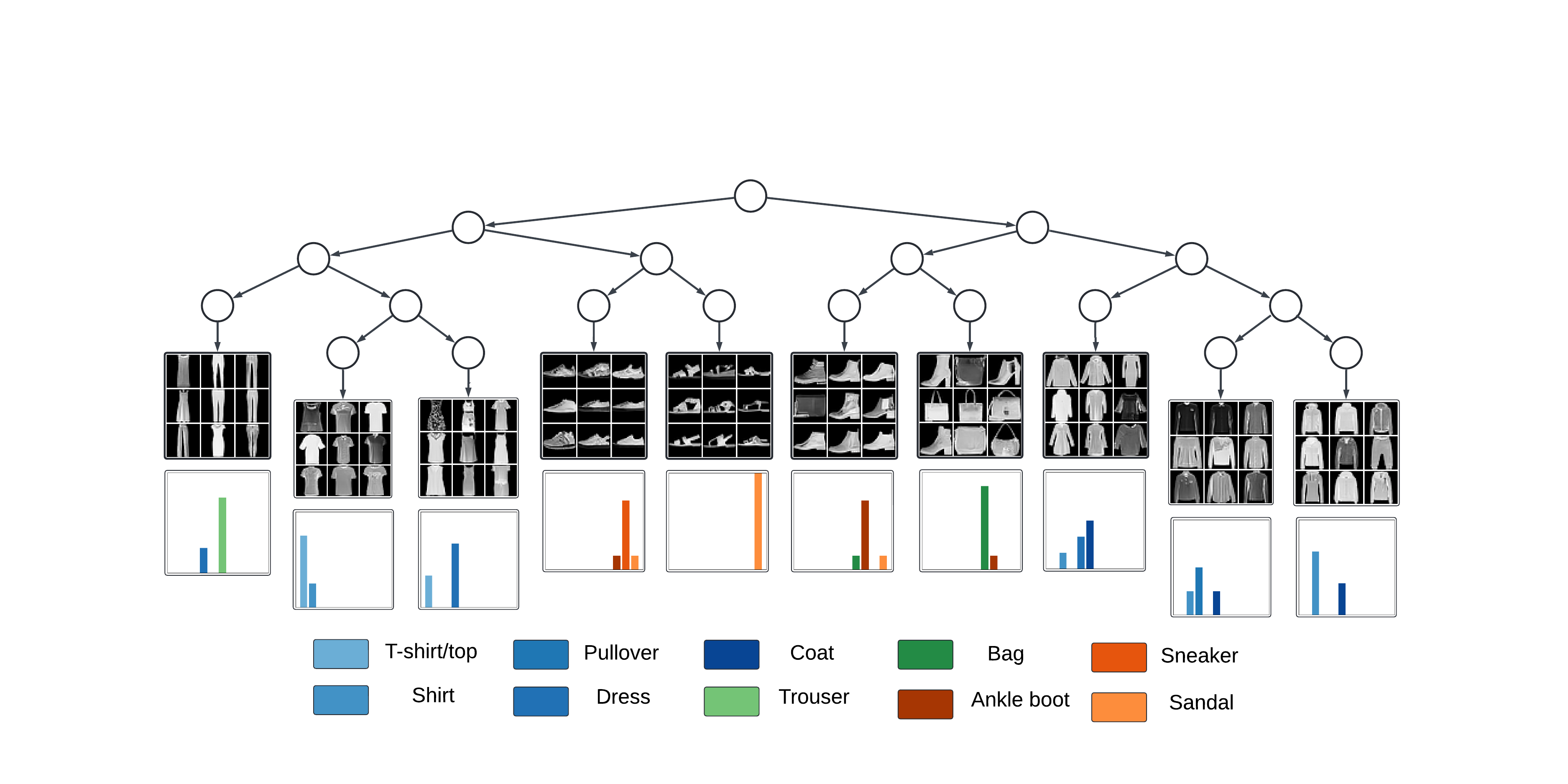}}
\caption{TreeDiffusion model trained on FashionMNIST. For each cluster, random newly generated images are displayed. Below each set of images, a normalized histogram (ranging from 0 to 1) shows the distribution of predicted classes from an independent, pre-trained classifier on FashionMNIST for all newly generated images in each leaf with a significant probability of reaching that leaf.}
\label{fmnist_generations_hist}
\end{center}
\end{figure*}

% \begin{figure*}[ht!]
% \begin{center}
% \centerline{\includegraphics[width=0.9\textwidth, trim=0 150 150 100, clip]{images/cifar_tree_generation_new.pdf}}
% \caption{TreeDiffusion model trained on CIFAR-10. For each cluster, random newly generated images are displayed. Below each set of images, a normalized histogram (ranging from 0 to 1) shows the distribution of predicted classes from an independent, pre-trained classifier on CIFAR-10 for all newly generated images in each leaf with a significant probability of reaching that leaf.}
% \label{cifar_generations_hist}
% \end{center}
% \end{figure*}

\begin{figure*}[ht!]
\begin{center}
\centerline{\includegraphics[width=\textwidth, trim=0 30 100 0, clip]{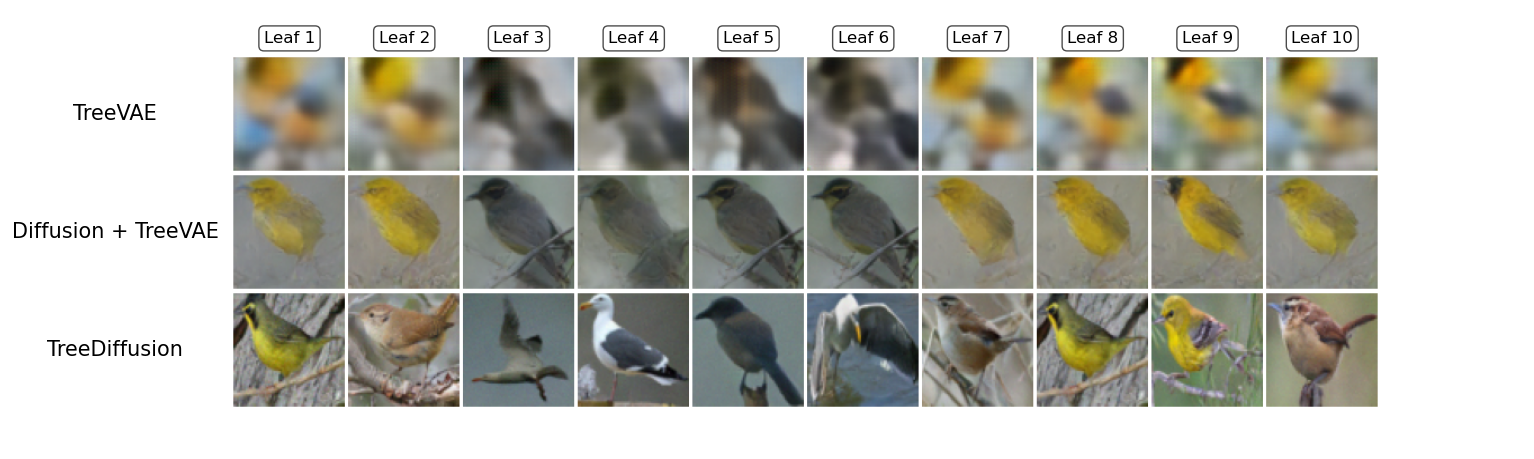}}
\caption{Image generations from each leaf of (top) TreeVAE, (middle) TreeVAE + Diffusion which starts denoising with the TreeVAE images, and (bottom) TreeDiffusion model conditioned on the hierarchical path embeddings, all trained on CUBICC. Each row displays the generated images from all leaves of the specified model, starting with the same sample from the root. The corresponding leaf probabilities are shown at the top of the image and are, by design, the same for all models.}
\label{cifar_generations_generations}
\end{center}
\end{figure*}

\subsubsection{Cluster information is retained across generations.} To quantitively assess whether the newly generated images retain their cluster information, we train a classifier on the original labeled training data and then use it to classify generated images of TreeDiffusion. Specifically, we classify the generations for each cluster separately. The idea is that ``pure`` leaves should create samples that are classified into one or very few classes. For this classification task, we use a ResNet-50 model \cite{he_deep_2016} trained on each dataset.
In \cref{fmnist_generations_hist}, we present randomly generated images from a TreeDiffusion model trained on FashionMNIST, together with normalized histograms depicting the distribution of the predicted classes for each leaf. For instance, clusters representing trousers and bags appear to accurately and distinctly capture their respective classes, as all their generated images are classified into one group only. Conversely, certain clusters are characterized by a mixture of classes, indicating that they are grouped together. Overall, we observe that the leaf-specific generations retain the hierarchical clustering structure found by TreeVAE, thereby enhancing the interpretability in diffusion models.

\subsubsection{On the benefits of hierarchical conditioning.} We hereby assess whether the conditioning on hierarchical representations improves cluster-specific generative quality. To this end, we compare the generations of TreeDiffusion, which is conditioned on the hierarchical representation, to the baseline TreeVAE + Diffusion from earlier, which is not conditioned on the latent cluster information. For this experiment, we use the previously introduced independent classifier to create the normalized histograms for each leaf to evaluate how cluster-specific the newly generated images are. As mentioned above, ideally, the majority of generated images from one leaf should be classified into one or very few classes from the original dataset. To quantify this, we compute the average entropy for all cluster-specific histograms. Lower entropy indicates less variation in the histograms and, thus, more cluster-specific generations. Table \ref{tab:cond_vs_uncond} presents the results for all labeled datasets. 

\begin{table}[ht] 
\centering
\caption{Cluster-specificity of TreeDiffusion generations comparing cluster-unconditional and cluster-conditional reverse models, measured by mean entropy. Lower entropy indicates more cluster-specific generations. The best result for each dataset is marked in \textbf{bold}.}
\label{tab:cond_vs_uncond}
\begin{tabular*}{\linewidth}{@{\extracolsep{\fill}}llcc}
\hline 
\textbf{Dataset} & \textbf{Method} & \textbf{Cluster Conditioning} & \textbf{Mean Entropy}\\
\hline 
MNIST    
& Diffusion + TreeVAE & $\times$ & $1.24$\\
& TreeDiffusion & $\checkmark$ & $\textbf{0.33}$\\
\hline 
Fashion     
& Diffusion + TreeVAE & $\times$ & $0.66$\\
& TreeDiffusion & $\checkmark$ & $\textbf{0.65}$\\
\hline 
CIFAR10  
& Diffusion + TreeVAE & $\times$ & $1.12$\\
& TreeDiffusion & $\checkmark$ & $\textbf{0.93}$\\
\hline 
CUBICC  
& Diffusion + TreeVAE & $\times$ & $\textbf{0.07}$\\
& TreeDiffusion & $\checkmark$ & $0.20$\\
\hline
\end{tabular*}
\end{table}

For most datasets, the conditional model exhibits lower mean entropy, indicating that cluster conditioning indeed helps guide the model to generate more distinct and representative images for each leaf. However, for the CUBICC dataset, we observe that the mean entropy is lower for the cluster-unconditional model. This is because the classifier tends to predict all images into a single class, a result of model degeneration, where it primarily generates images for only a few classes. \cref{cifar_generations_generations} visually presents the leaf generations for one sample of these models alongside the underlying TreeVAE generations. It can be observed that both the cluster-unconditional and conditional models exhibit a significant improvement in image quality. However, the images in the cluster-conditional model are more diverse, demonstrating greater adaptability for each cluster. Notably, across all models, the leaf-specific images share common properties, such as background color and overall shape, sampled at the root while varying in cluster-specific features from leaf to leaf within each model.

% shorten caption, table as floating box on side of the text, better margin

% \begin{table}[ht]
% \centering
% \begin{minipage}{0.5\linewidth} % Adjust to half the text width
% \caption{Cluster-specificity of TreeDiffusion generations for cluster-unconditional and cluster-conditional reverse models, measured by mean entropy. Lower entropy indicates more cluster-specific generations. Mean entropy is computed across all leaf-specific histograms of the predicted classes for newly generated images. The best result for each dataset is marked in \textbf{bold}.}
% \label{tab:cond_vs_uncond}

%     \begin{tabular*}{\linewidth}{@{\extracolsep{\fill}}llc}
%     \hline 
%     \textbf{Dataset} & \textbf{Method} & \textbf{Mean Entropy}\\
%     \hline MNIST    
%     & unconditional       & $1.24$\\
%     & conditional         & $\textbf{0.13}$\\
%     \hline Fashion     
%     & unconditional       & $\textbf{0.66}$\\
%     & conditional         & $\textbf{0.66}$\\
%     \hline CIFAR10  
%     & unconditional       & $1.12$\\
%     & conditional         & $\textbf{0.82}$\\
%     \hline
%     \end{tabular*}
% \end{minipage}
% \end{table}

\subsection{Ablation study on conditioning information} \label{ablation}

Finally, we conduct an ablation study to evaluate the impact of different conditioning signals on the generative performance. Specifically, we compare three types of conditioning: (i) hierarchical clustering signals derived from the latent embeddings of the selected cluster path $\boldsymbol{z}_{\mathcal{P}_l}$, (ii) flat clustering signals, including leaf assignment $l$, leaf embedding $\boldsymbol{z}_l$, or both, and (iii) an unconditioned setting where the diffusion model does not utilize any latent cluster representations from TreeVAE. Additionally, we examine the effect of using the TreeVAE leaf reconstruction $\hat{\vx}_{0}^{(l)}$ as the starting point for the denoising process in the second-stage diffusion model. The results, outlined in Table \ref{tab:ablation_study}, show the FID score calculated from $10,000$ samples generated using $100$ DDIM steps, averaged over $10$ seeds. Note that the first row in the table represents the TreeVAE + Diffusion model from the previous experiments, whereas the last row corresponds to the proposed TreeDiffusion method.

\begin{comment}
    
\begin{table}[ht]
\centering
\caption{Effect of conditioning signals on generative performance for CIFAR-10. FID scores for $10,000$ samples (lower is better) computed across $10$ random model initializations.}
\label{tab:ablation_study}
\begin{tabular}{cccc|cc}
    \toprule
    \textbf{Leaf Reconstruction} & \textbf{Leaf Assignment} & \textbf{Leaf Embedding} & \textbf{Path} & \textbf{FID} $\downarrow$ \\
    \textbf{$\hat{\vx}_{0}^{(l)}$} & \textbf{$l$} & \textbf{$\boldsymbol{z}_l$} & \textbf{$\boldsymbol{z}_{\mathcal{P}_l}$} & \\
    \midrule
    \cmark & \xmark & \xmark & \xmark & $19.7 \,\text{{\footnotesize $\pm\, 0.2$}}$ \\
    \cmark & \cmark & \xmark & \xmark & $19.1 \,\text{{\footnotesize $\pm\, 0.3$}}$ \\
    \cmark & \xmark & \cmark & \xmark & $18.9 \,\text{{\footnotesize $\pm\, 0.3$}}$ \\
    \cmark & \cmark & \cmark & \xmark & $19.2 \,\text{{\footnotesize $\pm\, 0.2$}}$ \\
    \xmark & \cmark & \cmark & \xmark & $19.1 \,\text{{\footnotesize $\pm\, 0.5$}}$ \\
    \midrule
    \cmark & \xmark & \xmark & \cmark & $18.2 \,\text{{\footnotesize $\pm\, 0.3$}}$ \\
    \xmark & \xmark & \xmark & \cmark & $\textbf{17.8} \,\text{{\footnotesize $\pm\, 0.4$}}$ \\
    %\midrule
    \midrule
    \multicolumn{4}{c|}{Vanilla DDIM} & $18.1 \,\text{{\footnotesize $\pm\, 0.3$}}$ \\
    \bottomrule
\end{tabular}
\end{table}

\end{comment}

\begin{table}[ht]
\centering
\caption{Effect of conditioning signals on generative performance for CIFAR-10. FID scores for $10,000$ samples (lower is better) computed across $10$ random model initializations.}
\label{tab:ablation_study}
\begin{tabular}{c|c|ccc|c}
    \toprule
    \textbf{Conditioning Type} & \quad \textbf{$\hat{\vx}_{0}^{(l)}$} \quad & \quad \textbf{$l$} & \quad \textbf{$\boldsymbol{z}_l$} & \quad \textbf{$\boldsymbol{z}_{\mathcal{P}_l}$} \quad & \quad \textbf{FID} $\downarrow$ \\
    \midrule
    No Cluster Conditioning & \cmark & \xmark & \xmark & \xmark & $19.7 \,\text{{\footnotesize $\pm\, 0.2$}}$ \\
    \midrule
    \multirow{4}{*}{Flat Clustering-Based Conditioning}  
        & \cmark & \cmark & \xmark & \xmark & $19.1 \,\text{{\footnotesize $\pm\, 0.3$}}$ \\
        & \cmark & \xmark & \cmark & \xmark & $18.9 \,\text{{\footnotesize $\pm\, 0.3$}}$ \\
        & \cmark & \cmark & \cmark & \xmark & $19.2 \,\text{{\footnotesize $\pm\, 0.2$}}$ \\
        & \xmark & \cmark & \cmark & \xmark & $19.1 \,\text{{\footnotesize $\pm\, 0.5$}}$ \\
    \midrule
    \multirow{2}{*}{Hierarchical Clustering-Based Conditioning \quad} 
        & \cmark & \xmark & \xmark & \cmark & $18.2 \,\text{{\footnotesize $\pm\, 0.3$}}$ \\
        & \xmark & \xmark & \xmark & \cmark & $\textbf{17.8} \,\text{{\footnotesize $\pm\, 0.4$}}$ \\
    % \midrule
    % \multicolumn{5}{c|}{Unconditional DDIM Baseline} & $18.1 \,\text{{\footnotesize $\pm\, 0.3$}}$ \\
    \bottomrule
\end{tabular}
\end{table}

The findings suggest that incorporating latent leaf information — whether through leaf assignment, leaf embedding, or both — significantly improves generative performance compared to relying solely on leaf reconstructions. This highlights the added benefit of conditioning on flat clustering information. Furthermore, conditioning on the full path $\boldsymbol{z}_{\mathcal{P}_l}$, which integrates all embeddings and intermediate node assignments from the root to the leaf, leads to an even greater performance boost. This underscores the effectiveness of hierarchical clustering information beyond flat clustering. As a result, harnessing $\boldsymbol{z}_{\mathcal{P}_l}$ from the hierarchical structure not only produces more structured generations, as illustrated in \cref{leaf_samples}, but also enhances the generative performance of generative clustering models. Notably, when conditioning on the full path, the model performs better without relying on TreeVAE reconstructions. Instead, the conditional diffusion model generates new images from scratch, guided solely by the latent information.

% Remarkably, conditioning on the latent path also exceeds the generative performance of a fully unconditional, vanilla DDIM model, which achieves an average FID of $18.1$. 

% -----------------------------------------------------------------------------

% \section{Discussion}

% -----------------------------------------------------------------------------

\section{Conclusion}
In this work, we present TreeDiffusion, a novel approach to integrate hierarchical clustering into diffusion models. 
By enhancing TreeVAE with a Denoising Diffusion Implicit Model conditioned on latent hierarchical representations, we propose a model capable of generating distinct, high-quality images that faithfully represent their respective data clusters. This approach not only improves the visual fidelity of the generated images but also facilitates cluster visualization. TreeDiffusion offers a robust framework that bridges the gap between clustering and generative performance, thereby expanding the potential applications of generative models in areas requiring detailed and more interpretable visual data interpretation.

\begin{credits}

\subsubsection{\ackname} 
Jorge da Silva Gonçalves is supported by the grant \#2021-911 of the Strategic Focal Area ``Personalized Health and Related Technologies (PHRT)'' of the ETH Domain (Swiss Federal Institutes of Technology). Laura Manduchi is supported by the SDSC PhD Fellowship \#1-001568-037. Moritz Vandenhirtz is supported by the Swiss State Secretariat for Education, Research and Innovation (SERI) under contract number MB22.00047.

\end{credits}

\newpage

% -----------------------------------------------------------------------------

% Of course, authors have complete freedom on how they choose to structure their paper. Section headers from Introduction up to and including Conclusions are completely up to the discretion of the authors; use whichever structure you see fit. Title, Abstract, the credits environment, and References, however, are mandatory.

\begin{comment}

\begin{credits}

\subsubsection{\ackname} Laura Manduchi is supported by the SDSC PhD Fellowship (number 1-001568-037). Moritz Vandenhirtz is supported by the Swiss State Secretariat for Education, Research and Innovation (SERI) under contract number MB22.00047.

% not sure if I have to include internships in this one?

\subsubsection{\discintname}
It is now necessary to declare any competing interests or to specifically
state that the authors have no competing interests. Please place the
statement with a bold run-in heading in small font size beneath the
(optional) acknowledgments,
for example: The authors have no competing interests to declare that are
relevant to the content of this article. Or: Author A has received research
grants from Company W. Author B has received a speaker honorarium from
Company X and owns stock in Company Y. Author C is a member of committee Z.
\end{credits}

\end{comment}

\bibliographystyle{splncs04}
\bibliography{references}
%% Note that this preceding line implies that you store your BibTeX references in a file called 'mybibliography.bib'. If you instead store your references in a file with a different name, for instance 'references.bib', the preceding line should read '\bibliography{references}'. Whatever you do, DO NOT put the file name extension .bib inside the \bibliography command; this will trip up LaTeX compilers. 
%
% If you do not want to use BibTeX, you can also type up the bibliography exactly as you see fit, using the following structure:
% % Note that this number 8 reserves an amount of space (equal to the natural width of the given number) for the label of your references; if you have more than 9 references, you will want to change this number to 18. If you have more than 19 references, this number is best changed to 88. If you have more than 99 references, I salute you.




% LNCS Homepage, \url{http://www.springer.com/lncs}, last accessed 2023/10/25
% \end{thebibliography}

\newpage

\appendix
\section*{Appendix}

\section{Detailed Formulation of TreeVAE}

\begin{figure}[!h]
  \centering
  \includegraphics[width=\textwidth]{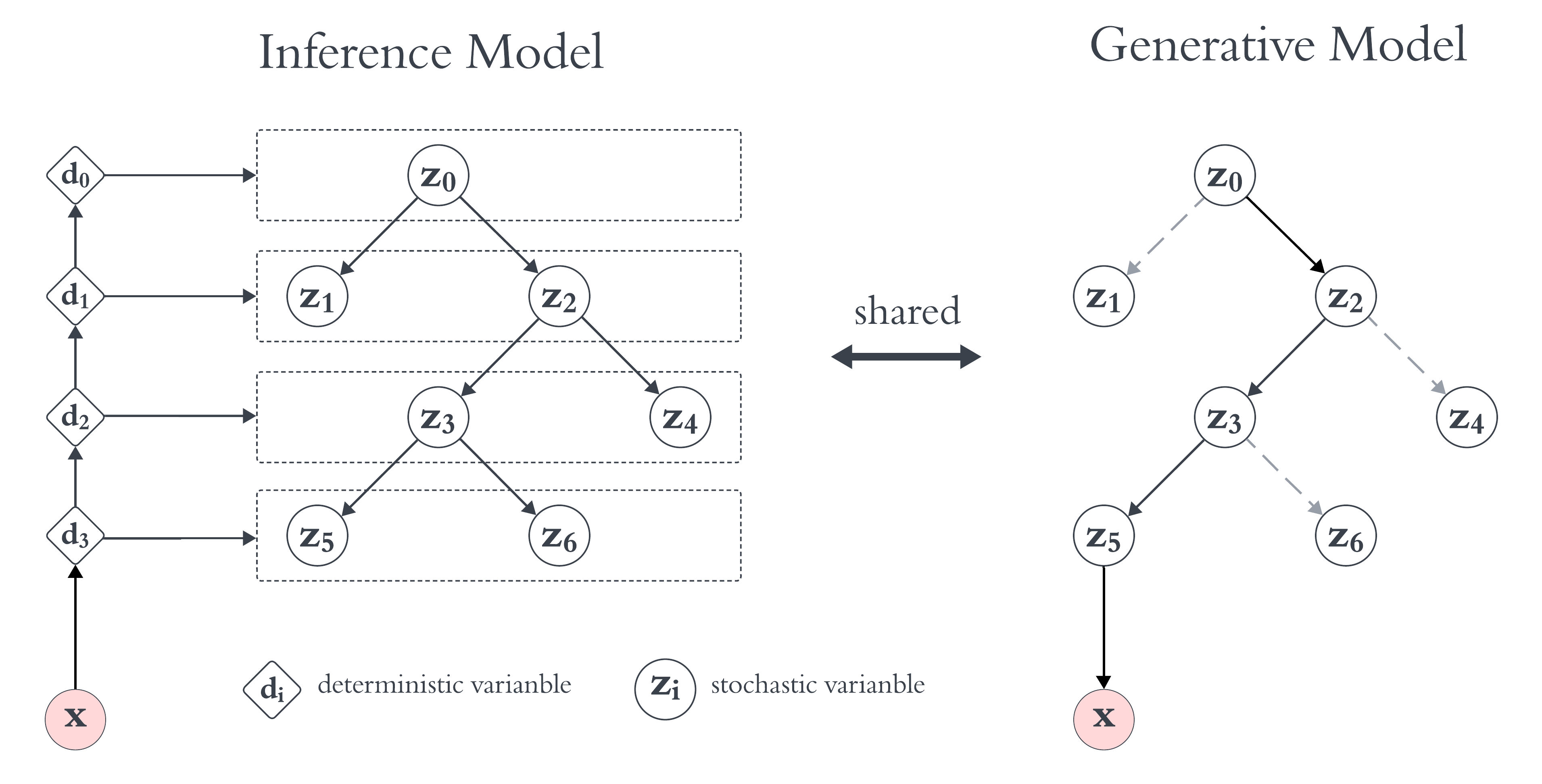}
  \caption[TreeVAE model structure]{Illustration of the TreeVAE model structure. The learned tree topology is shared between the inference and generative models.} \label{fig:treevae}
\end{figure}

Here, we provide a more in-depth introduction to TreeVAE, closely following the original paper \cite{manduchi_tree_2023}. Built upon the VAE framework \cite{kingma_auto-encoding_2013}, TreeVAEs inherently possess stochastic latent variables. However, unlike traditional VAEs, TreeVAEs organize these latent variables in a learnable binary tree structure as shown in Figure \ref{fig:treevae}, enabling them to capture complex hierarchical relationships. 

In the TreeVAE framework, the tree hierarchically divides the data, yielding separate stochastic embeddings for each node. This division into nodes induces a probability distribution, allowing each sample to navigate through the nodes of the tree in a probabilistic manner, from the root to the leaf. Ideally, high-variance features should be partitioned at earlier stages in the tree, while deeper nodes encapsulate more detailed concepts. The flexible tree structure is learned during training and is specific to the data distribution. This structure allows for the incorporation of sample-specific probability distributions over the different paths in the tree, as further explained in Section \ref{sec:model_formulation}. In this manner, TreeVAEs contribute to the generation of comprehensive hierarchical data representations.

TreeVAE combines elements from both LadderVAE \cite{sonderby_ladder_2016} and ANT \cite{tanno_adaptive_2019}, with the aim to create a VAE-based model capable of performing hierarchical clustering via the latent variables. Similar to LadderVAE, the inference and the generative model share the same top-down hierarchical structure. In fact, if we isolate a path from the root to any leaf in the tree structure of TreeVAE, we obtain an instance of the top-down model seen in LadderVAE. On the other side, both Adaptive Neural Trees and TreeVAE engage in representation and architecture learning during training. However, while ANTs are tailored for regression and classification tasks, TreeVAE's focus lies in hierarchical clustering and generative modeling. This allows TreeVAEs to generate new class-specific data samples based on the latent embeddings of the leaves from the tree.

\subsection{Model Formulation} \label{sec:model_formulation}

%As depicted in Figure \ref{fig:treevae}, 
TreeVAE comprises an inference model and a generative model, which share the global structure of the binary tree $\mathcal{T}$. Following the original work \cite{manduchi_tree_2023}, the tree structure is learned during training using dataset $\boldsymbol{X}$ and a predetermined maximum tree depth denoted as $H$. Specifically, the tree structure entails the set of nodes $\mathbb{V} = \{0,...,V\}$, the subset of leaves $\mathbb{L} \subset \mathbb{V}$, and the set of edges $\mathcal{E}$. 

While the global structure of the tree is the same for all samples in the data, the latent embeddings $\boldsymbol{z} = \{\boldsymbol{z}_0,...,\boldsymbol{z}_V\}$ for all nodes and the so-called decisions $\boldsymbol{c} = \{\mathrm{c}_0,...,\mathrm{c}_{V-\mid \mathbb{L} \mid}\}$ for all non-leaf nodes are unique to each sample, representing sample-specific random variables. The latent embeddings $\boldsymbol{z}$ are modeled as Gaussian random variables. Their distribution parameters are determined by functions that depend on the latent embeddings of the parent node. These functions, referred to as ``transformations'', are implemented as MLP.  Moreover, each decision variable $\mathrm{c}_i$ corresponds to a Bernoulli random variable, where $\mathrm{c}_i = 0$ signifies the selection of the left child at internal node $i$. These decision variables influence the traversal path within the tree during both the generative and inference processes. The parameters governing the Bernoulli distributions are functions of the respective node value, parametrized by MLP termed ``routers''. 

The transformations and routers are learned during the model training process, as further elaborated in Section \ref{model_training_treevae}, which comprehensively covers all aspects of model training. The exact parametrizations of the transformations and routers vary depending on whether they are applied in the context of the inference or generative model. Subsequently, the specifics of both of these models will be explored.

\subsection*{Generative Model} \label{inf_model_treevae}

The top-down generative model of TreeVAE, illustrated in Figure \ref{fig:treevae} on the right side, governs the generation of new samples. It starts by sampling a latent representation $\boldsymbol{z}_0$ from a standard Gaussian distribution for the root node, serving as the initial point for the generation process:
\begin{equation}
p_\theta\left(\boldsymbol{z}_0\right)=\mathcal{N}\left(\boldsymbol{z}_0 \mid \mathbf{0}, \boldsymbol{I}\right) .
\end{equation}
This latent embedding then traverses through the tree structure, leading to new samples being generated at each node based on their ancestor nodes. These remaining nodes in the tree are characterized by their own latent representations, which are sampled conditionally on the sample-specific embeddings of their parent nodes. This process can be mathematically expressed as follows:
\begin{equation}
p\left(\boldsymbol{z}_i \mid \boldsymbol{z}_{pa(i)}\right)=
\mathcal{N}\left(\boldsymbol{z}_i \mid \mu_{p, i}(\boldsymbol{z}_{pa(i)}), \sigma_{p, i}^2(\boldsymbol{z}_{pa(i)})\right), 
\end{equation}
where $\left\{\mu_{p, i}, \sigma_{p, i} \mid i \in \mathbb{V} \backslash\{0\}\right\}$ correspond to the transformation neural networks associated with the generative model, as denoted by the subscript $p$.  

Following the generation of latent representations for each node $i$ in the tree, the decision regarding traversal to the left or right child node are determined based on the sampled $\boldsymbol{z}_i$ value. These decisions, denoted by $\mathrm{c}_i$, for each non-leaf node $i$ are guided by routers, which are neural network functions linked to the generative model and defined as $\left\{r_{p, i} \mid i \in \mathbb{V} \backslash \mathbb{L}\right\}$. Here, $\mathrm{c}_i = 0$ indicates the selection of the left child of the internal node $i$. As mentioned before, these decisions follow a Bernoulli distribution,
$\mathrm{c}_i \mid \boldsymbol{z}_i \sim \operatorname{Ber}\left(r_{p, i}\left(\boldsymbol{z}_i\right)\right),$
which, in turn, induces the following path probability between a node $i$ and its parent node $pa(i)$: 
\begin{equation}
p\left(\mathrm{c}_{pa(i) \rightarrow i} \mid \boldsymbol{z}_{pa(i)}\right) =
\operatorname{Ber}\left(\mathrm{c}_{pa(i) \rightarrow i} \mid r_{p,{pa(i)}} (\boldsymbol{z}_{pa(i)})\right).
\end{equation}
Each decision made at a parent node determines the selection of the subsequent child node along the path. Consequently, the generative process progresses recursively until a leaf node $l$ is reached. Importantly, each decision path $\mathcal{P}_l$ from the root to the leaf $l$  is associated with a distinct set of decisions, which, in turn, determine the probabilities of traversing to different child nodes. These path probabilities, computed by the routers, play a critical role in shaping the distribution of the generated samples. The path $\mathcal{P}_l$ can be defined by the set of traversed nodes in the path. Moreover, let $\boldsymbol{z}_{\mathcal{P}_l}=\left\{\boldsymbol{z}_i \mid i \in \mathcal{P}_l\right\}$ denote the set of latent embeddings for each node in the path $\mathcal{P}_l$. Then, the prior probability of the latent embeddings and the path given the tree structure $\mathcal{T}$ corresponds to
\begin{equation}
    p_\theta\left(\boldsymbol{z}_{\mathcal{P}_l}, \mathcal{P}_l\right)=p\left(\boldsymbol{z}_0\right) \prod_{i \in \mathcal{P}_l \backslash\{0\}} \underbrace{p\left(\mathrm{c}_{p a(i) \rightarrow i} \mid \boldsymbol{z}_{p a(i)}\right)}_{\text {decision probability}} \underbrace{p\left(\boldsymbol{z}_i \mid \boldsymbol{z}_{p a(i)}\right)}_{\text {sample probability}}.
\end{equation}
To conclude the generative process, $\boldsymbol{x}$ is obtained based on the latent embeddings of a selected leaf $l$. Nevertheless, assumptions on the distribution of the inputs are required. For real-valued $\boldsymbol{x} $, such as in colored datasets, \cite{manduchi_tree_2023} assume a Gaussian distribution. For grayscale images, they consider the Bernoulli distribution. Thus, 
\begin{equation}
p_\theta\left(\boldsymbol{x} \mid \boldsymbol{z}_{\mathcal{P}_l}, \mathcal{P}_l\right)
= 
    \begin{cases}
      \mathcal{N}\Big(\boldsymbol{x} \mid \mu_{x, l}\left(\boldsymbol{z}_l\right), \sigma_{x, l}^2\left(\boldsymbol{z}_l\right)\Big)  & \text{for colored datasets,} \\
      \operatorname{Ber}\Big(\boldsymbol{x} \mid \mu_{x, l}\left(\boldsymbol{z}_l\right)\Big)& \text{for grayscale datasets,}
    \end{cases}
\end{equation}
where $\left\{\mu_{x, l}, \sigma_{x, l} \mid l \in \mathbb{L}\right\}$ or $\left\{\mu_{x, l} \mid l \in \mathbb{L}\right\}$, respectively, are implemented as leaf-specific neural networks called ``decoders". Typically, a simplification is made by assuming $\sigma_{x, l}^2\left(\boldsymbol{z}_l\right) = \boldsymbol{I}$ for convenience.

\subsection*{Inference Model} \label{gen_model_treevae}

The inference model of TreeVAE, depicted on the left side of Figure \ref{fig:treevae}, introduces a deterministic bottom-up pass to incorporate the conditioning on $\boldsymbol{x}$, distinguishing it from the generative model. This bottom-up process is akin to the framework proposed by LadderVAE \cite{sonderby_ladder_2016}. In this hierarchical structure, the bottom-up deterministic variables depend on each other via a series of neural network operations, specifically MLP of the same architecture as the transformation MLP defined previously,
\begin{equation}
\begin{aligned} \label{eq:bottom-up_MLP}
\mathbf{d}_h & =\operatorname{MLP}\left(\mathbf{d}_{h+1}\right) .
\end{aligned}
\end{equation}
The first deterministic variable in this chain, denoted as $\mathbf{d}_H$, serves as the output of an encoder neural network. This encoder accepts the input image $\boldsymbol{x}$ and transforms it into the flattened, low-dimensional vector $\mathbf{d}_H$. $H$ corresponds to both the number of deterministic variables involved in the bottom-up process and the maximum depth to which the tree structure can grow during training as further explained in Section \ref{model_training_treevae}.

The tree structure is shared between the inference and the generative model, though adjustments are made to the node-specific parameterizations of the Gaussian distributions. Notably, in the inference phase, information is introduced through conditioning on $\boldsymbol{x}$, influencing the distribution of latent embeddings across all nodes. Hence, similar to LadderVAE, the means $\boldsymbol{\mu}_{q, i}$ and variances $\boldsymbol{\sigma}_{q, i}^2$ of the variational posterior distribution for each node $i$ are calculated using dense linear network layers conditioned on the deterministic variable of the same depth as the node $i$: 
\begin{align}
\hat{\boldsymbol{\mu}}_{q, i} & = \operatorname{Linear}\left(\mathbf{d}_{\operatorname{depth}(i)}\right), \quad i \in \mathbb{V} \label{eq:dense_mu} \\
\hat{\boldsymbol{\sigma}}_{q, i}^2 & = \operatorname{Softplus}\left(\operatorname{Linear}\left(\mathbf{d}_{\operatorname{depth}(i)}\right)\right), \quad i \in \mathbb{V} \label{eq:dense_sig} \\
\boldsymbol{\sigma}_{q, i}&=\frac{1}{\hat{\boldsymbol{\sigma}}_{q, i}^{-2}+\boldsymbol{\sigma}_{p, i}^{-2}}, \quad \boldsymbol{\mu}_{q, i}=\frac{\hat{\boldsymbol{\mu}}_{q, i} \hat{\boldsymbol{\sigma}}_{q, i}^{-2}+\boldsymbol{\mu}_{p, i} \boldsymbol{\sigma}_{p, i}^{-2}}{\hat{\boldsymbol{\sigma}}_{q, i}^{-2}+\boldsymbol{\sigma}_{p, i}^{-2}},
\end{align}
where the subscript $q$ denotes the parameters specific to the inference model. Given these posterior parameters, the following equations determine the variational distributions of the latent embeddings for the root and the succeeding nodes in the tree structure:
\begin{align}
q\left(\boldsymbol{z}_0 \mid \boldsymbol{x}\right)&=\mathcal{N}\left(\boldsymbol{z}_0 \mid \mu_{q, 0}(\boldsymbol{x}), \sigma_{q, 0}^2(\boldsymbol{x})\right), \label{eq:var_root} \\
q_\phi\left(\boldsymbol{z}_i \mid \boldsymbol{z}_{p a(i)}\right)&=\mathcal{N}\left(\boldsymbol{z}_i \mid \mu_{q, i}\left(\boldsymbol{z}_{p a(i)}\right), \sigma_{q, i}^2\left(\boldsymbol{z}_{pa(i)}\right)\right), \quad \forall i \in \mathcal{P}_l. \label{eq:var_nodes}
\end{align}
The routers in the inference model maintain the same architecture as those in the generative model. Nevertheless, in the inference process, decisions are now conditioned on $\boldsymbol{x}$. This means that while the routers retain their original structure, the distribution of decision variables $\mathrm{c}_i$ at node $i$ now depends on the deterministic variable of the corresponding depth, $\mathrm{c}_i \mid \boldsymbol{x} \sim 
\operatorname{Ber}(r_{q, i}(\mathbf{d}_{\operatorname{depth}(\mathrm{i})})),
$ resulting in the following variational path probability between a node $i$ and its parent node $pa(i)$: 
\begin{equation} \label{eq:var_decisions}
q\left(\mathrm{c}_{pa(i) \rightarrow i} \mid \boldsymbol{x}\right) =q\left(\mathrm{c}_i \mid \mathbf{d}_{\operatorname{depth}(pa(i))}\right)=
\operatorname{Ber}\left(\mathrm{c}_{pa(i) \rightarrow i} \mid r_{q, pa(i)}\left(\mathbf{d}_{\operatorname{depth}(pa(i))}\right)\right), 
\end{equation}

Finally, given the variational distributions of the latent embeddings \eqref{eq:var_root} \& \eqref{eq:var_nodes} and the variational distributions of the decisions \eqref{eq:var_decisions}, the variational posterior distribution of the latent embeddings and paths corresponds to:
\begin{equation}
q\left(\boldsymbol{z}_{\mathcal{P}_l}, \mathcal{P}_l \mid \boldsymbol{x}\right)=q\left(\boldsymbol{z}_0 \mid \boldsymbol{x}\right) \prod_{i \in \mathcal{P}_l \backslash\{0\}} q\left(\mathrm{c}_{p a(i) \rightarrow i} \mid \boldsymbol{x}\right) q\left(\boldsymbol{z}_i \mid \boldsymbol{z}_{p a(i)}\right) .
\end{equation}

\subsection{Model Training} \label{model_training_treevae}

TreeVAE \cite{manduchi_tree_2023} entails the training of various components, involving both the parameters of the neural network layers present in the inference and generative models, as detailed in Section \ref{sec:model_formulation}, and the binary tree structure $\mathcal{T}$. The training process alternates between optimizing the model parameters with a fixed tree structure and expanding the tree.

During a training iteration given the current tree structure, the neural layer parameters are optimized. These include the parameters of the inference model and the generative model, encompassing the encoder ($\mu_{q, 0}, \sigma_{q, 0}$), the bottom-up MLPs from \eqref{eq:bottom-up_MLP}, the dense linear layers from \eqref{eq:dense_mu} \& \eqref{eq:dense_sig}, the transformations $\left(\left\{\left(\mu_{p, i}, \sigma_{p, i}\right),\left(\mu_{q, i}, \sigma_{q, i}\right) \mid i \in \mathbb{V} \backslash\{0\}\right\}\right)$, the routers $\left(\left\{r_{p, i}, r_{q, i} \mid i \in \mathbb{V} \backslash \mathbb{L}\right\}\right)$, and the decoders $\left(\left\{\mu_{x, l}, \sigma_{x, l} \mid l \in \mathbb{L}\right\}\right)$. The objective in training these parameters is to maximize the likelihood of the data given the learned tree structure $\mathcal{T}$, denoted as $p(\boldsymbol{x} \mid \mathcal{T})$. This corresponds to modeling the distribution of real data via the hierarchical latent embeddings. To compute $p(\boldsymbol{x} \mid \mathcal{T})$, the latent embeddings must be marginalized out from the joint distribution:
\begin{equation}
p(\boldsymbol{x} \mid \mathcal{T})=\sum_{l \in \mathbb{L}} \int_{\boldsymbol{z}_{\mathcal{P}_l}} p\left(\boldsymbol{x}, \boldsymbol{z}_{\mathcal{P}_l}, \mathcal{P}_l\right)=\sum_{l \in \mathbb{L}} \int_{\boldsymbol{z}_{\mathcal{P}_l}} p_\theta\left(\boldsymbol{z}_{\mathcal{P}_l}, \mathcal{P}_l\right) p_\theta\left(\boldsymbol{x} \mid \boldsymbol{z}_{\mathcal{P}_l}, \mathcal{P}_l\right).
\end{equation}
Similar to other Variational Autoencoders \cite{kingma_auto-encoding_2013}, the optimization of TreeVAE ultimately comes down to maximizing the ELBO. As shown by \cite{manduchi_tree_2023}, the ELBO in this setting can be written as follows:
\begin{equation} \label{eq:elbo_treevaes}
\mathcal{L}(\boldsymbol{x} \mid \mathcal{T}):= 
\underbrace{\mathbb{E}_{q(\boldsymbol{z}_{\mathcal{P}_l}, \mathcal{P}_l \mid \boldsymbol{x})}[\log p(\boldsymbol{x} \mid \boldsymbol{z}_{\mathcal{P}_l}, \mathcal{P}_l)]}_{\mathclap{\displaystyle \mathcal{L}_{rec}}} - 
\underbrace{\operatorname{KL}(q(\boldsymbol{z}_{\mathcal{P}_l}, \mathcal{P}_l \mid \boldsymbol{x}) \| p(\boldsymbol{z}_{\mathcal{P}_l}, \mathcal{P}_l))_{
\phantom{q(\boldsymbol{z}_{\mathcal{P}_l}, \mathcal{P}_l \mid \boldsymbol{x})}
\hspace{-45pt}}}_{\mathclap{\displaystyle \operatorname{KL}_{root} + \operatorname{KL}_{nodes} + \operatorname{KL}_{decisions}}}.
\end{equation}
Hereby, we distinguish between the reconstruction term and the KL divergence term between the variational posterior and the prior of the tree, which can be further decomposed into contributions from the root, the remaining nodes, and decisions, as indicated in Equation \eqref{eq:elbo_treevaes}. Both terms are approximated using Monte Carlo (MC) sampling during training, as elaborated further in \cite{manduchi_tree_2023}.

To grow the binary tree structure $\mathcal{T}$, TreeVAE begins with a simple tree configuration, typically composed of a root and two leaves. This initial structure is trained for a defined number of epochs, optimizing the ELBO. Subsequently, the model iteratively expands the tree by attaching two new child nodes to a current leaf node in the model. In their approach, the authors \cite{manduchi_tree_2023} opted to expand the tree by selecting nodes with the highest number of assigned samples, thus implicitly encouraging balanced leaves. The sub-tree formed by the new leaves and the parent node undergoes training for another number of epochs, keeping the weights of the remaining model frozen. This expansion process continues until either the tree reaches its maximum depth $H$, a predefined maximum number of effective leaves, or another predefined condition is met. Optionally, after the tree has been expanded, all parameters in the model may be fine-tuned for another predefined number of epochs. Finally, the tree is pruned to remove empty branches.

Contrastive learning is used to improve the clustering performance, especially for colored images. This addition enables the model to encode prior knowledge on data invariances through augmentations, facilitating the learning process and better capturing meaningful relationships within complex data. By incorporating contrastive objectives into the training process, TreeVAE becomes more adept at retrieving semantically meaningful clusters from colored image data.

\section{Detailed Formulation of Diffusion Models} \label{sec:diffusion}

Diffusion models have garnered considerable attention in recent years for their remarkable image-generation capabilities, outperforming traditional GANs in achieving high-quality results. This surge in interest has led to the development of numerous diffusion-based models, with the initial concept being introduced by \cite{sohl-dickstein_deep_2015}, drawing inspiration from principles rooted in thermodynamics. In this section, we delve into one of the first and most well-known diffusion models, namely the DDPM, outlined in Section \ref{sec:DDPM}. Additionally, we explore one possible integration of DDPM with VAEs to leverage the interpretable latent space offered by VAEs along with the superior generation quality of DDPM. Note that we use the DDIM instantiation at inference time.

\subsection{Diffusion Denoising Probabilistic Models} \label{sec:DDPM}

Diffusion Denoising Probabilistic Models (DDPM) \cite{ho_denoising_2020} are latent variable models that consist of two opposed processes. The main idea behind DDPM involves iteratively adding noise to the original image, thereby progressively degrading the signal with each step until only noise remains. In a second step, the image is successively reconstructed through a denoising process, employing a learned function to remove the noise. Therefore, DDPM essentially entail a forward noising process followed by a reverse denoising process, aiming to restore the original image from its noisy counterpart. Subsequently, both these processes will be explored in more detail.

\subsection*{Forward Process}

The forward process, also known as the diffusion process or forward noising process, serves a similar purpose as the inference model in VAEs. It operates as a Markov chain that gradually introduces noise to the data signal $\boldsymbol{x}_0$ over $T$ steps, where the intermediate states of the deformed data are denoted as $\boldsymbol{x}_t$ for $t = 1, \dots, T$. This process follows a trajectory determined by a noise schedule ${ \beta_1, \dots, \beta_T }$, which controls the rate at which the original data is degraded. While it is possible to learn the variances $\beta_t$ via reparametrization, they are often chosen as hyperparameters with a predetermined schedule \cite{ho_denoising_2020}. Assuming Gaussian noise, the forward process can be represented as follows:
\begin{align}
q\left(\boldsymbol{x}_{1: T} \mid \boldsymbol{x}_0\right)&=\prod_{t=1}^T q\left(\boldsymbol{x}_t \mid \boldsymbol{x}_{t-1}\right) \\
q\left(\boldsymbol{x}_t \mid \boldsymbol{x}_{t-1}\right)&=\mathcal{N}\left(\sqrt{1-\beta_t} \boldsymbol{x}_{t-1}, \beta_t \boldsymbol{I}\right)
\end{align}

Importantly, $\boldsymbol{x}_t$ for any step $t$ can be sampled directly given $\boldsymbol{x}_0$ using:
\begin{equation}
    q\left(\boldsymbol{x}_t \mid \boldsymbol{x}_0\right)=\mathcal{N}\left(\sqrt{\bar{\alpha}_t} \boldsymbol{x}_0,\left(1-\overline{\alpha}_t\right) \boldsymbol{I}\right),
\end{equation}
where $\alpha_t=\left(1-\beta_t\right)$ and $\bar{\alpha}_t=\prod_{s=1}^t \alpha_s$. This property significantly boosts training efficiency by eliminating the need to sample every state between $\boldsymbol{x}_0$ and $\boldsymbol{x}_t$. Additionally, when conditioned on $\boldsymbol{x}_0$, the posteriors of the forward process are tractable and can be determined in closed form:
\begin{align} \label{eq:ddpm_forward_post}
q\left(\boldsymbol{x}_{t-1} \mid \boldsymbol{x}_t, \boldsymbol{x}_0\right) & =\mathcal{N}\left(\boldsymbol{x}_{t-1} ; \tilde{\boldsymbol{\mu}}_t\left(\boldsymbol{x}_t, \boldsymbol{x}_0\right), \tilde{\beta}_t \boldsymbol{I}\right), \\ \label{eq:noisy_mean}
\text { where } \quad \tilde{\boldsymbol{\mu}}_t\left(\boldsymbol{x}_t, \boldsymbol{x}_0\right) & =\frac{\sqrt{\bar{\alpha}_{t-1}} \beta_t}{1-\bar{\alpha}_t} \boldsymbol{x}_0+\frac{\sqrt{\alpha_t}\left(1-\bar{\alpha}_{t-1}\right)}{1-\bar{\alpha}_t} \boldsymbol{x}_t \quad  
\text { and } \quad \tilde{\beta}_t =\frac{1-\bar{\alpha}_{t-1}}{1-\bar{\alpha}_t} \beta_t.
\end{align}

\subsection*{Reverse Process}

The reverse process, also known as the reverse denoising process or backward process, is comparable to the generative model in Variational Autoencoders. Beginning with pure Gaussian noise, the reverse process governs the sample generation by progressively eliminating noise through a sequence of $T$ denoising steps. These steps are facilitated by time-dependent, learnable Gaussian transitions that adhere to the Markov property, meaning each transition relies solely on the state of the previous time step in the Markov chain. This denoising sequence reverses the noise addition steps of the forward process, enabling the gradual recovery of the original data. 

In summary, the reverse process models a complex target data distribution by sequentially transforming simple distributions through a generative Markov chain. This approach overcomes the traditional tradeoff between tractability and flexibility in probabilistic models \cite{sohl-dickstein_deep_2015}, leading to a flexible and efficient generative model: 
\begin{align} \label{eq:ddpm_reverse_post}
    p\left(\boldsymbol{x}_{0: T}\right)&=p\left(\boldsymbol{x}_T\right) \prod_{t=1}^T p_\theta\left(\boldsymbol{x}_{t-1} \mid \boldsymbol{x}_t\right) \\
    p_\theta\left(\boldsymbol{x}_{t-1} \mid \boldsymbol{x}_t\right)&=\mathcal{N}\left(\boldsymbol{\mu}_\theta\left(\boldsymbol{x}_t, t\right), \boldsymbol{\Sigma}_\theta\left(\boldsymbol{x}_t, t\right)\right)
\end{align}
\cite{ho_denoising_2020} keep the time dependent variances $\boldsymbol{\Sigma}_\theta\left(\boldsymbol{x}_t, t\right)=\sigma_t^2 \boldsymbol{I}$ constant, typically choosing $\sigma_t^2=\beta_t$ or $\sigma_t^2=\tilde{\beta}_t=\frac{1-\bar{\alpha}_{t-1}}{1-\bar{\alpha}_t} \beta_t$. Therefore, only the time-dependent posterior mean function $\boldsymbol{\mu}_\theta$ is learned during training. In fact, \cite{ho_denoising_2020} further suggest the following reparametrization of the posterior means: 
\begin{align*}
\boldsymbol{\mu}_\theta\left(\boldsymbol{x}_t, t\right)&=\tilde{\boldsymbol{\mu}}_t\left(\boldsymbol{x}_t, \frac{1}{\sqrt{\bar{\alpha}_t}}\left(\boldsymbol{x}_t-\sqrt{1-\bar{\alpha}_t} \boldsymbol{\epsilon}_\theta\left(\boldsymbol{x}_t\right)\right)\right) \\ &=\frac{1}{\sqrt{\alpha_t}}\left(\boldsymbol{x}_t-\frac{\beta_t}{\sqrt{1-\bar{\alpha}_t}} \boldsymbol{\epsilon}_\theta\left(\boldsymbol{x}_t, t\right)\right),
\end{align*}
where $\boldsymbol{\epsilon}_\theta$ is a learnable function that predicts the noise at any given time step $t$.

\subsection*{Model Training}

Training a diffusion model corresponds to optimizing the reverse Markov transitions to maximize the likelihood of the data. This is achieved by maximizing the ELBO or, equivalently, minimizing the variational upper bound on the negative log-likelihood: 
\begin{align*}
\mathcal{L} = &\mathbb{E}_q[\underbrace{\operatorname{KL}\left(q\left(\boldsymbol{x}_T \mid \boldsymbol{x}_0\right) \| p\left(\boldsymbol{x}_T\right)\right)}_{\mathcal{L}_T} \\ 
&+\sum_{t>1} \underbrace{\operatorname{KL}\left(q\left(\boldsymbol{x}_{t-1} \mid \boldsymbol{x}_t, \boldsymbol{x}_0\right) \| p_\theta\left(\boldsymbol{x}_{t-1} \mid \boldsymbol{x}_t\right)\right)}_{\mathcal{L}_{t-1}} \\
&-\underbrace{\log p_\theta\left(\boldsymbol{x}_0 \mid \boldsymbol{x}_1\right)}_{\mathcal{L}_0}]
\end{align*}
For small diffusion rates $\beta_t$, the forward and the reverse processes share the same functional form, following Gaussian distributions \cite{sohl-dickstein_deep_2015}. Thus, the KL divergence terms between the posteriors of the forward process \eqref{eq:ddpm_forward_post} and the reverse process \eqref{eq:ddpm_reverse_post} have a closed form. The training process can be further simplified \cite{ho_denoising_2020}  by introducing a more efficient training objective. Given the assumption of fixed variances, $\mathcal{L}_{t-1}$ can be written as: 
\begin{equation} \label{eq:ddpm_obj1}
\mathcal{L}_{t-1}=\mathbb{E}_q\left[\frac{1}{2 \sigma_t^2}\left\|\tilde{\boldsymbol{\mu}}_t\left(\boldsymbol{x}_t, \boldsymbol{x}_0\right)-\boldsymbol{\mu}_\theta\left(\boldsymbol{x}_t, t\right)\right\|^2\right]+C, 
\end{equation}
where $C$ is a constant that does not depend on the model parameters $\theta$. Thus, the reverse process mean function, $\boldsymbol{\mu}_\theta$, is optimized to predict $\tilde{\boldsymbol{\mu}}_t$, the fixed noisy mean function at time step $t$ from the forward process \eqref{eq:noisy_mean}. However, instead of directly comparing $\boldsymbol{\mu}_\theta$ and $\tilde{\boldsymbol{\mu}}_t$, by reparametrization, the model can be trained to predict the noise $\boldsymbol{\epsilon}$ at any given time step $t$ which results in more stable training results according to \cite{ho_denoising_2020}. Thus, \eqref{eq:ddpm_obj1} can be further rewritten as 
\begin{equation} \label{eq:ddpm_objective}
    \mathcal{L}_{t-1} - C = \mathbb{E}_{\boldsymbol{x}_0, \boldsymbol{\epsilon}}\left[\frac{\beta_t^2}{2 \sigma_t^2 \alpha_t\left(1-\bar{\alpha}_t\right)}\left\|\boldsymbol{\epsilon}-\boldsymbol{\epsilon}_\theta\left(\sqrt{\bar{\alpha}_t} \boldsymbol{x}_0+\sqrt{1-\bar{\alpha}_t} \boldsymbol{\epsilon}, t\right)\right\|^2\right], 
\end{equation}
where $\boldsymbol{\epsilon}_\theta$ is a function parametrized as a neural network that maintains equal dimensionality for input and output. The preferred architecture for $\boldsymbol{\epsilon}_\theta$ is a U-Net \cite{ronneberger_u-net_2015}, as used by \cite{ho_denoising_2020,dhariwal_diffusion_2021,pandey_diffusevae_2022}. The training of $\boldsymbol{\epsilon}_\theta$ involves multiple epochs, where for each sample $\boldsymbol{x}_0$ of the original data, a time step $t$ within the diffusion sequence $1, \dots, T$ is chosen at random. Subsequently, some noise $\boldsymbol{\epsilon} \sim \mathcal{N}(\mathbf{0}, \boldsymbol{I})$ is sampled for that specific time step, which needs to be predicted by $\boldsymbol{\epsilon}_\theta$. To optimize the model, a gradient step is computed with respect to the loss function detailed in Equation \eqref{eq:ddpm_objective}. Therefore, this training procedure does not require iterating through every step of the diffusion model, circumventing the issue of slow sample generation that is typical for diffusion models.

\section{Additional Qualitative Results}

\subsection{Generations on CIFAR-10} 

\begin{figure*}[ht!]
\begin{center}
\centerline{\includegraphics[width=0.9\textwidth, trim=0 150 150 100, clip]{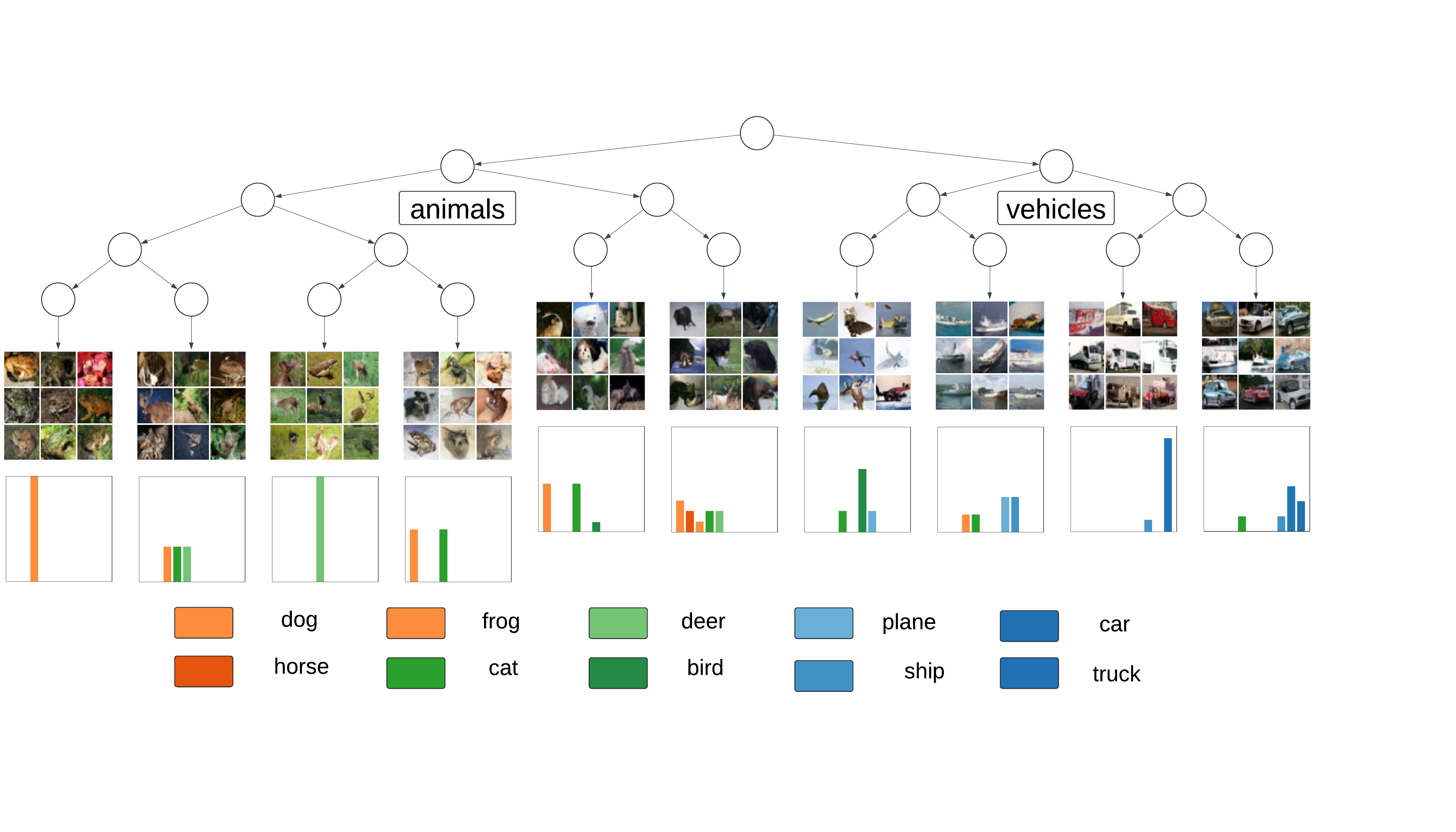}}
\caption{TreeDiffusion model trained on CIFAR-10. For each cluster, random newly generated images are displayed. Below each set of images, a normalized histogram (ranging from 0 to 1) shows the distribution of predicted classes from an independent, pre-trained classifier on MNIST for all newly generated images in each leaf with a significant probability of reaching that leaf.}
\label{cifar_generations_hist}
\end{center}
\end{figure*}

\cref{cifar_generations_hist} presents an additional plot similar to the one for FashionMNIST from the main text. This plot illustrates the generated images of the TreeDiffusion model when trained on the CIFAR-10 dataset. In each of these plots, we display randomly generated images for each cluster. Below each set of leaf-specific images, we provide a normalized histogram showing the distribution of predicted classes by an independent ResNet-50 classifier that has been pre-trained on the training data of the dataset. This visualization helps to understand how well the model can generate distinct and meaningful clusters in the context of different datasets.

\clearpage

\subsection{Additional Generation Examples for TreeVAE vs. TreeDiffusion} \label{further_gen_examples}

\begin{figure*}[ht!]
\begin{center}
\centerline{\includegraphics[width=\textwidth, trim=0 20 90 0, clip]{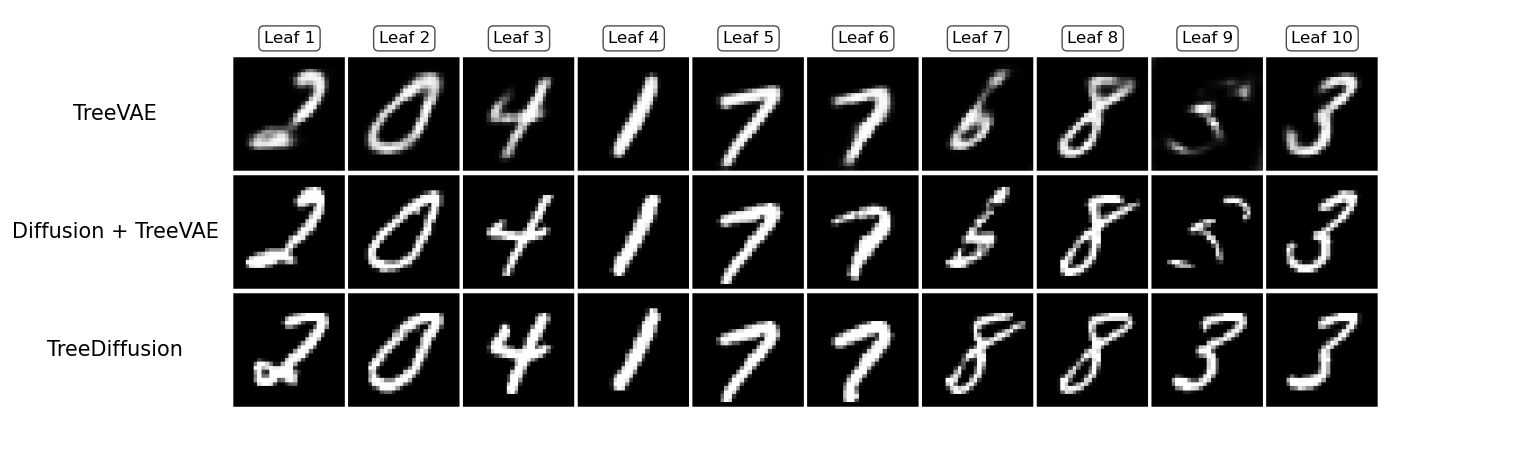}}
\centerline{\includegraphics[width=\textwidth, trim=0 20 90 0, clip]{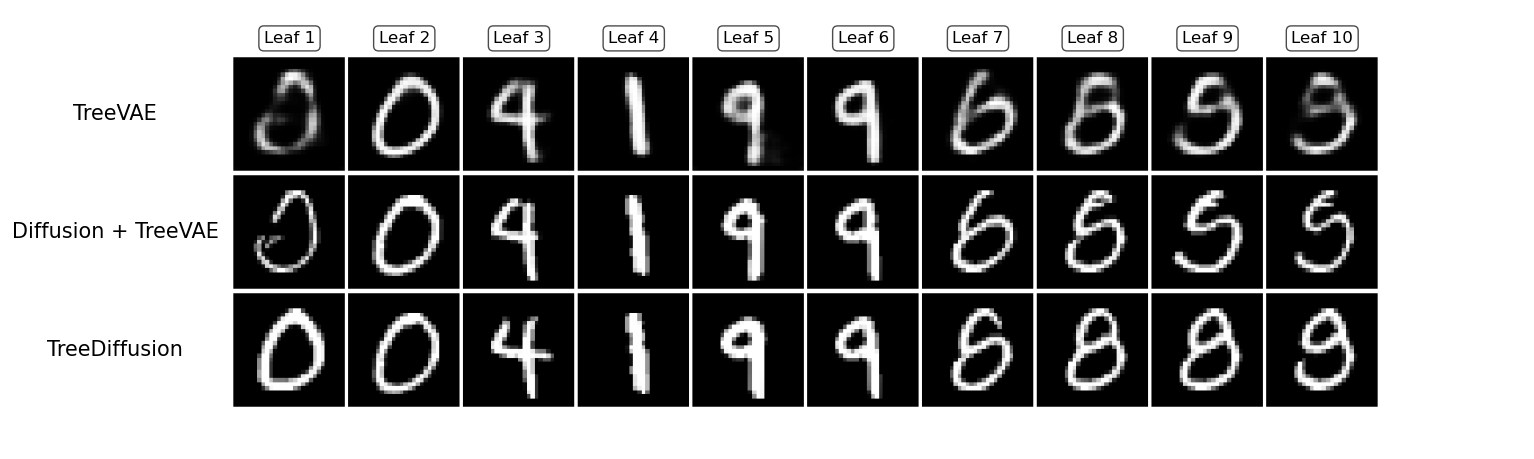}}
\centerline{\includegraphics[width=\textwidth, trim=0 20 90 0, clip]{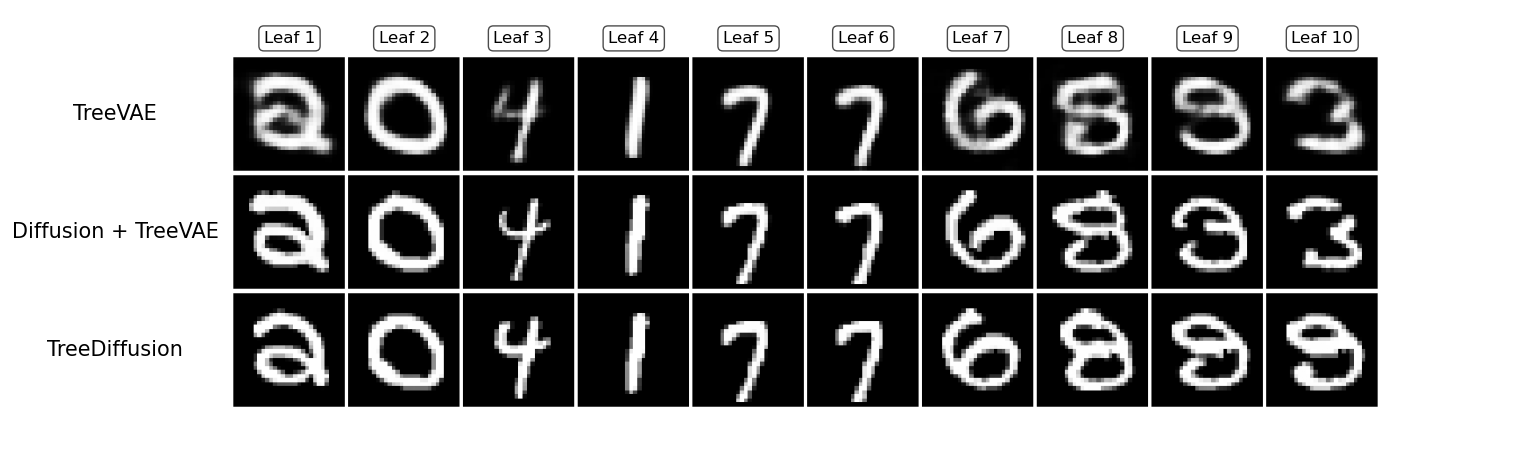}}
\centerline{\includegraphics[width=\textwidth, trim=0 20 90 0, clip]{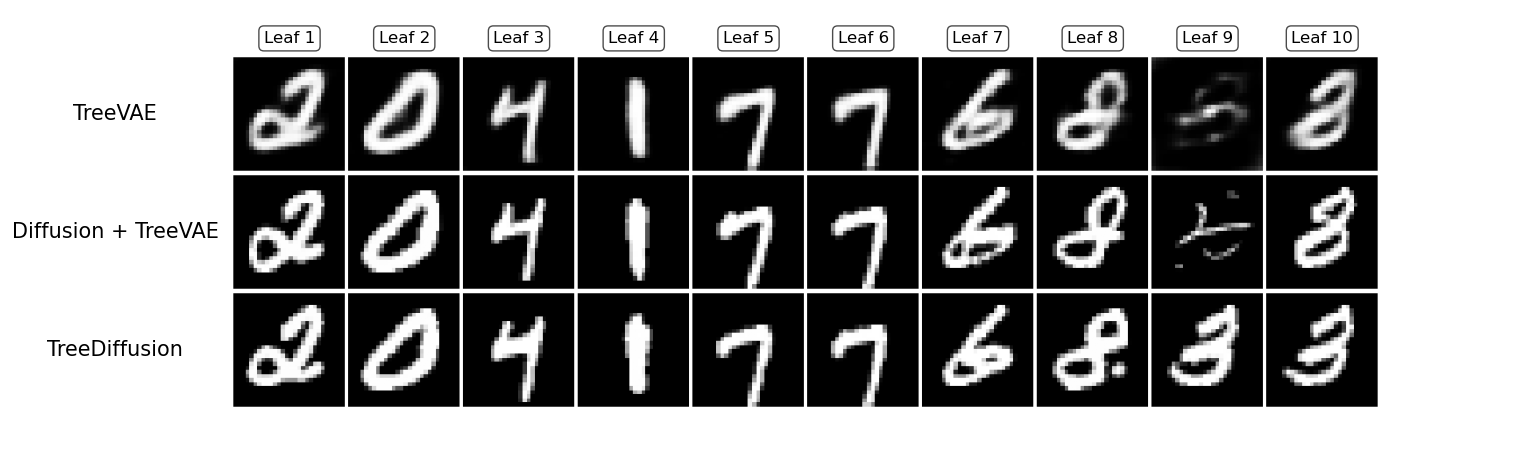}}
\caption{For each example, we show image generations from every leaf of TreeVAE, TreeVAE + Diffusion, and TreeDiffusion
model, all trained on MNIST. Each row shows the generated images from all leaves
of the respective model, starting with the same root sample.}
\label{mnist_generations_generations1}
\end{center}
\end{figure*}

\begin{figure*}[ht!]
\begin{center}
\centerline{\includegraphics[width=\textwidth, trim=0 20 90 0, clip]{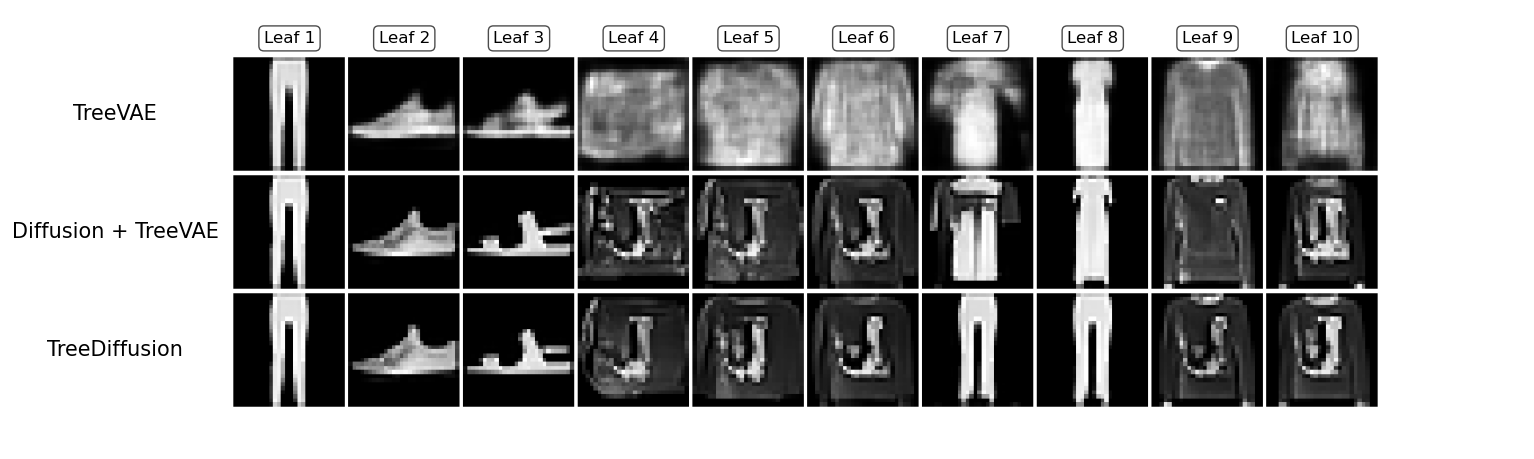}}
\centerline{\includegraphics[width=\textwidth, trim=0 20 90 0, clip]{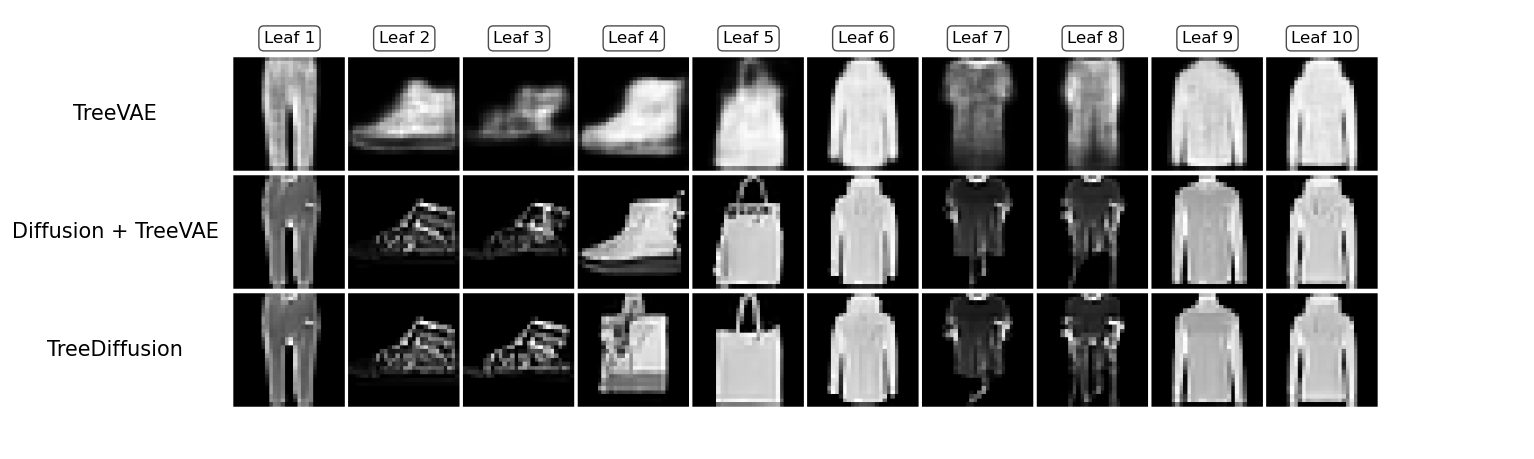}}
\centerline{\includegraphics[width=\textwidth, trim=0 20 90 0, clip]{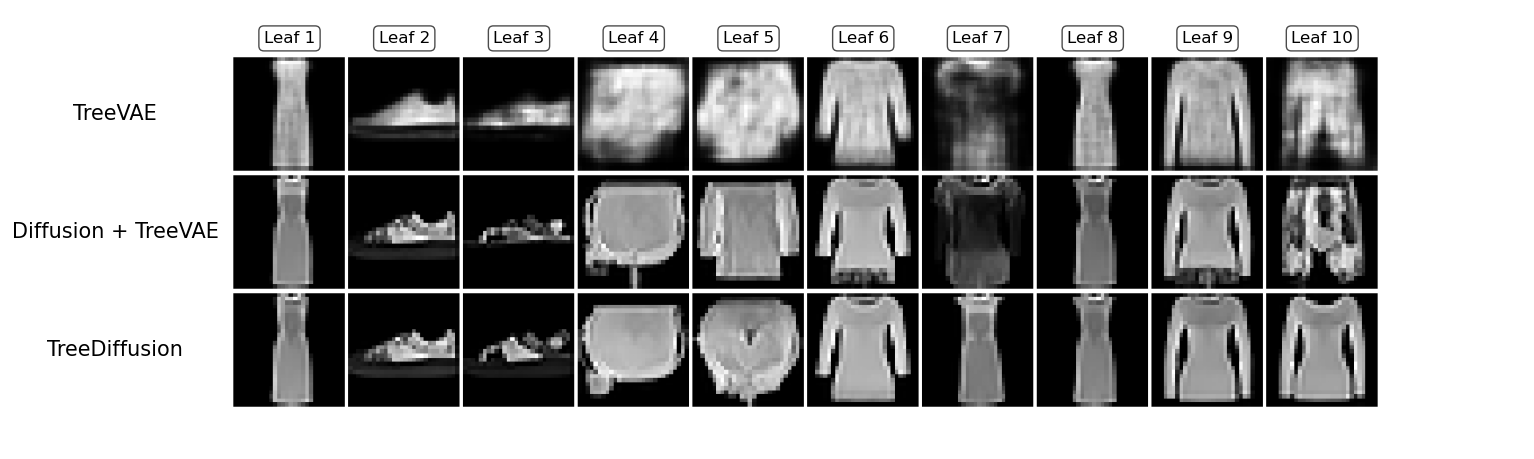}}
\centerline{\includegraphics[width=\textwidth, trim=0 20 90 0, clip]{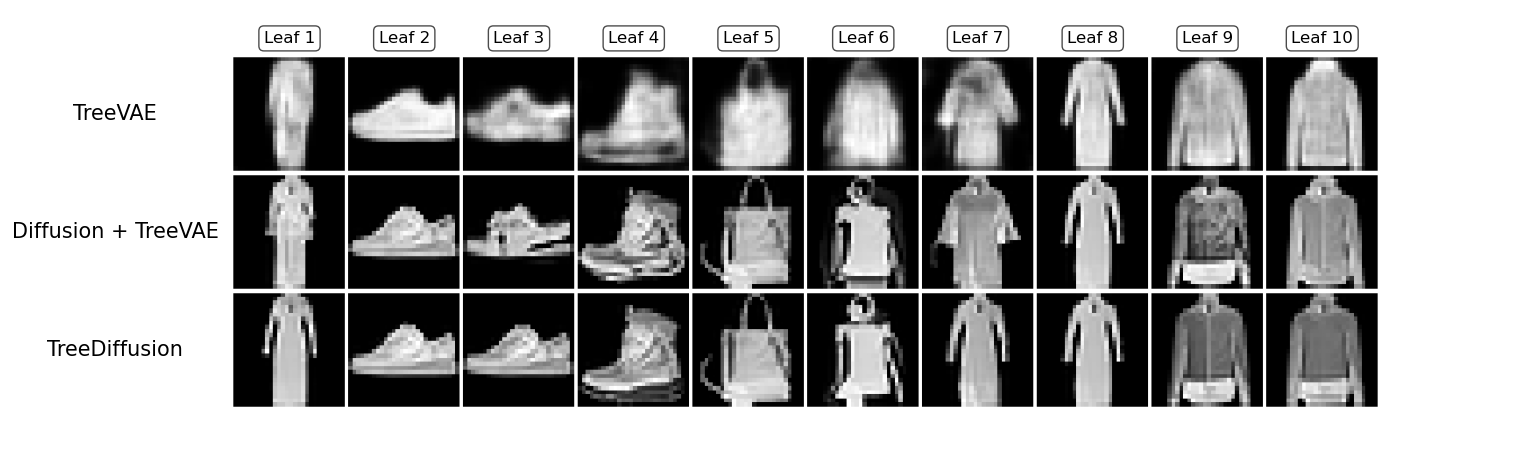}}
\caption{For each example, we show image generations from every leaf of TreeVAE, TreeVAE + Diffusion, and TreeDiffusion
model, all trained on FashionMNIST. Each row shows the generated images from all leaves
of the respective model, starting with the same root sample.}
\label{fmnist_generations_generations1}
\end{center}
\end{figure*}

\begin{figure*}[ht!]
\begin{center}
\centerline{\includegraphics[width=\textwidth, trim=0 20 90 0, clip]{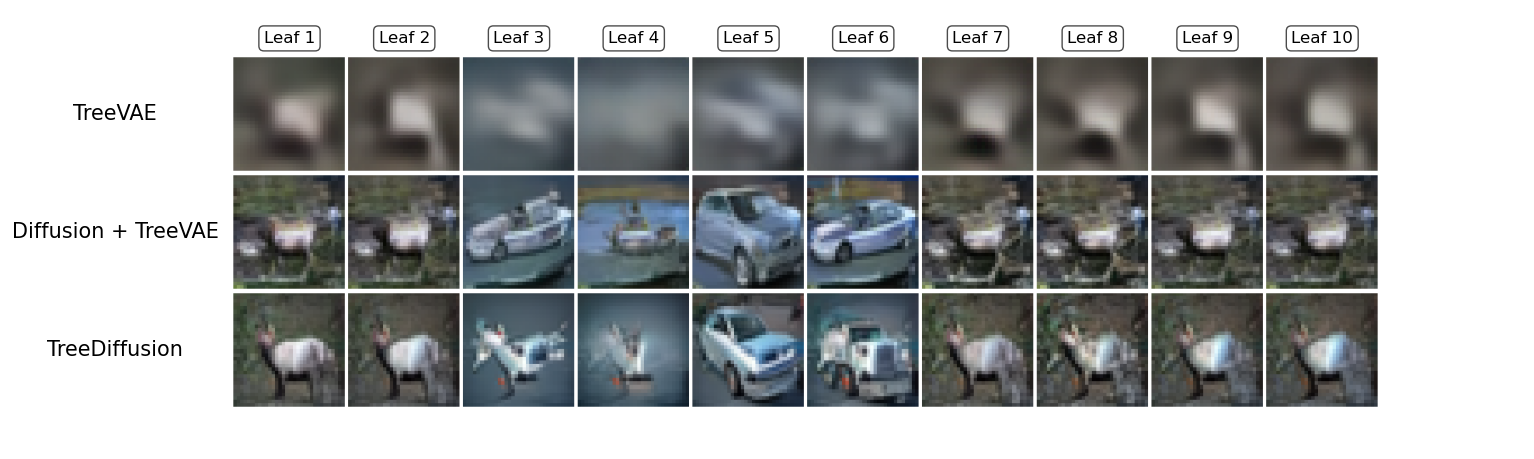}}
\centerline{\includegraphics[width=\textwidth, trim=0 20 90 0, clip]{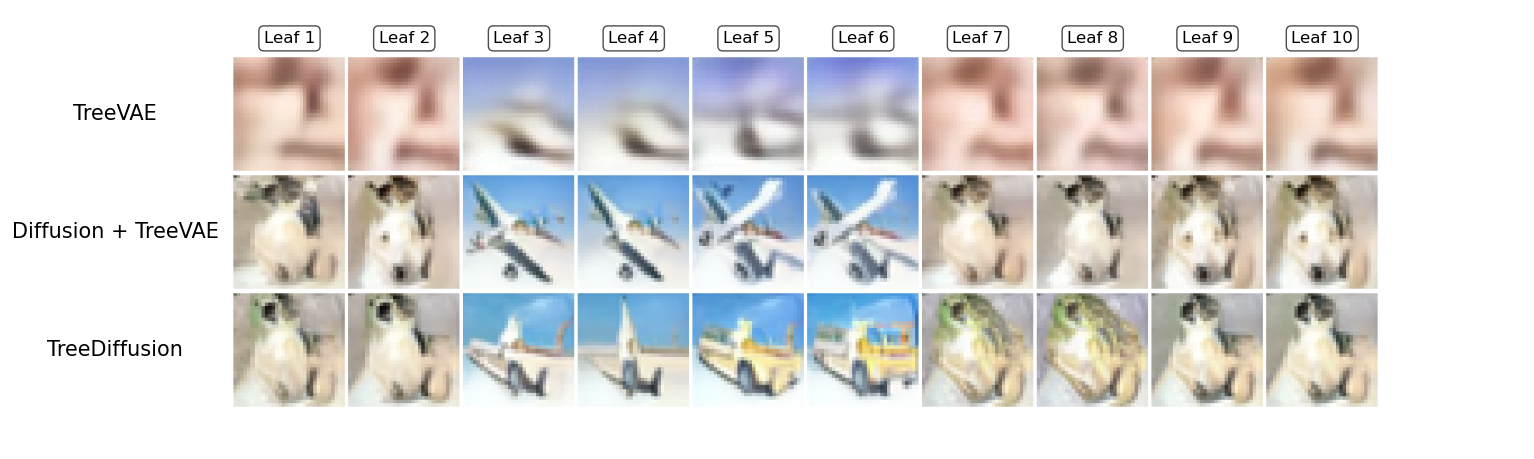}}
\centerline{\includegraphics[width=\textwidth, trim=0 20 90 0, clip]{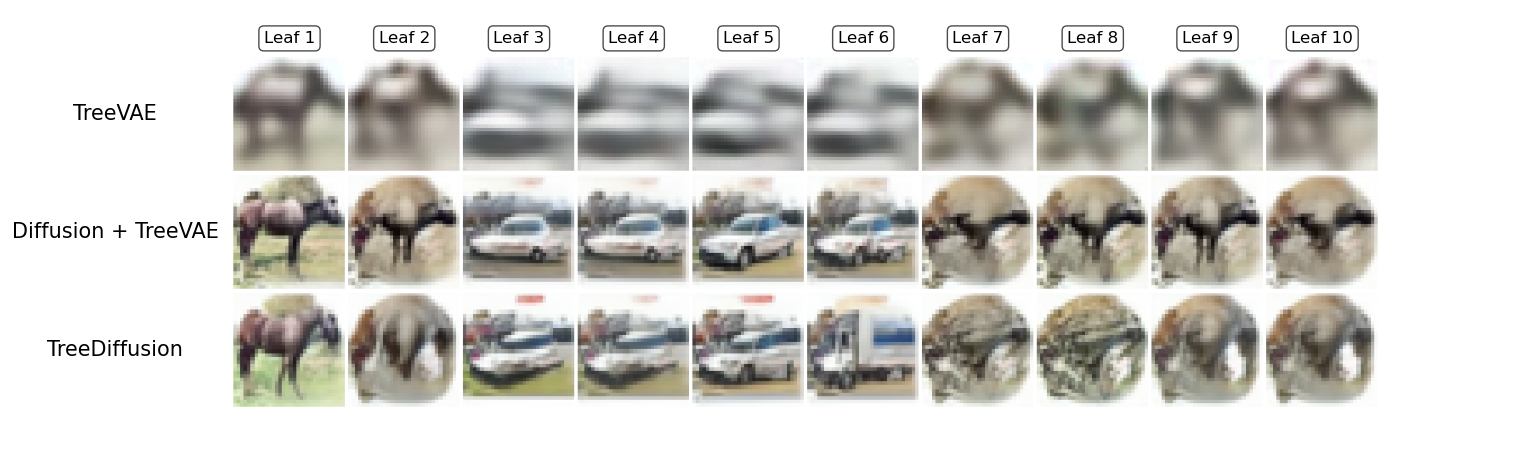}}
\centerline{\includegraphics[width=\textwidth, trim=0 20 90 0, clip]{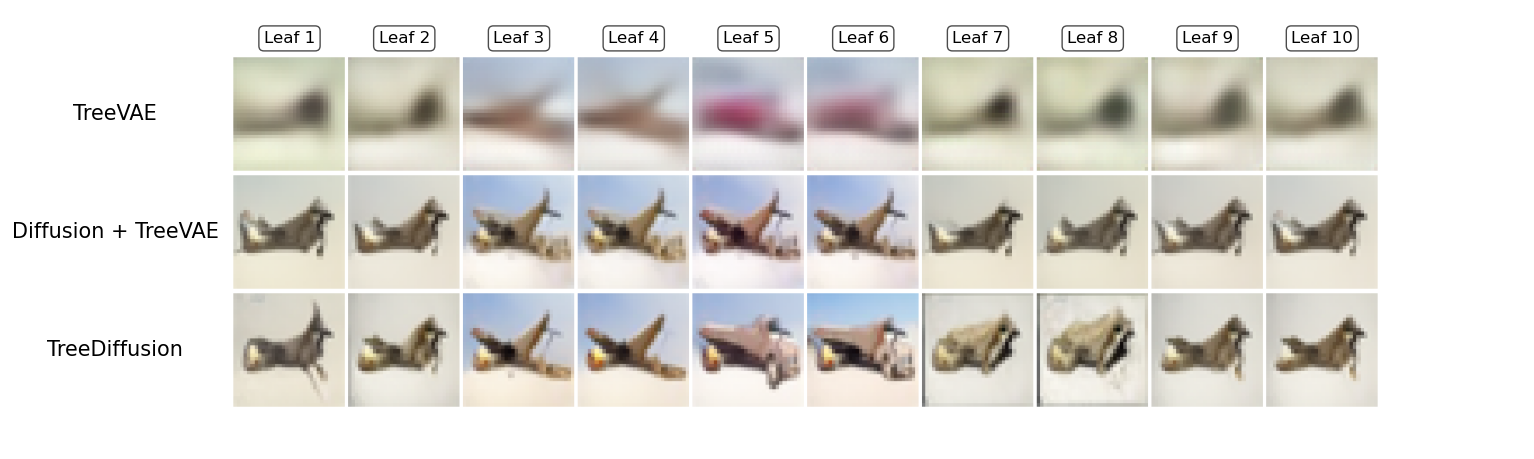}}
\caption{For each example, we show image generations from every leaf of TreeVAE, TreeVAE + Diffusion, and TreeDiffusion
model, all trained on CIFAR-10. Each row shows the generated images from all leaves
of the respective model, starting with the same root sample.}
\label{cifar_generations_generations1}
\end{center}
\end{figure*}

\begin{figure*}[ht!]
\begin{center}
\centerline{\includegraphics[width=\textwidth, trim=0 20 90 0, clip]{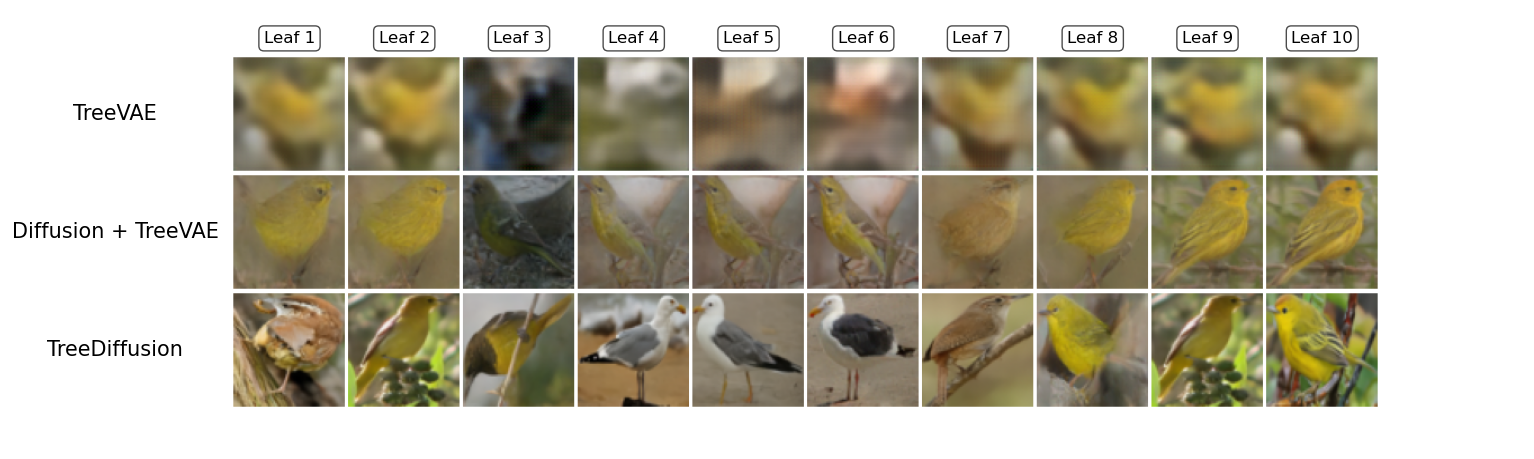}}
\centerline{\includegraphics[width=\textwidth, trim=0 20 90 0, clip]{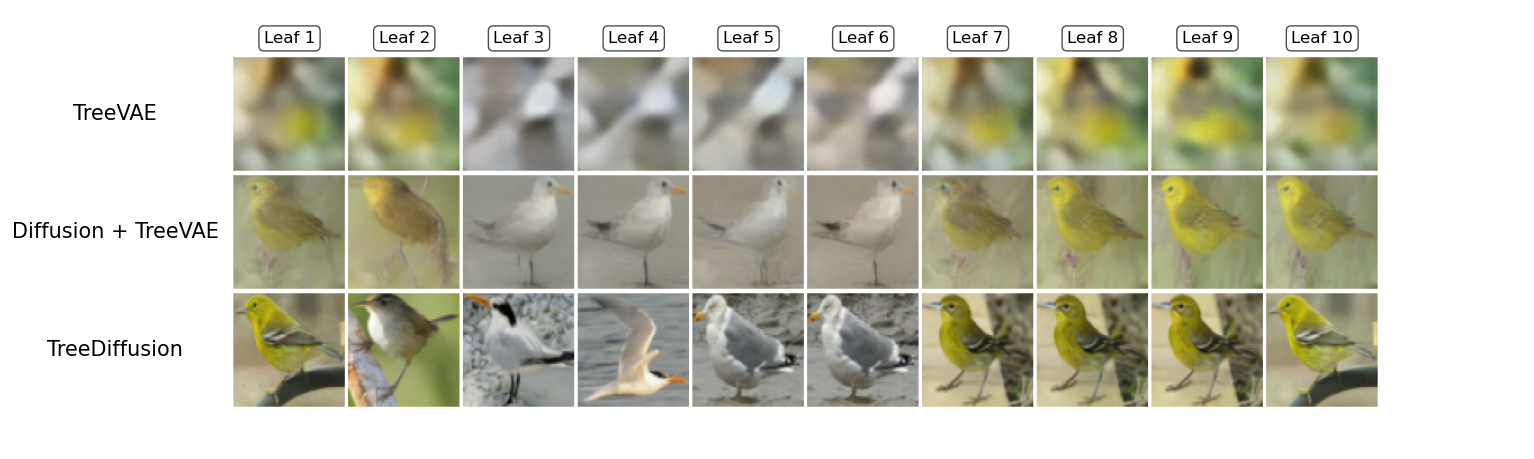}}
\centerline{\includegraphics[width=\textwidth, trim=0 20 90 0, clip]{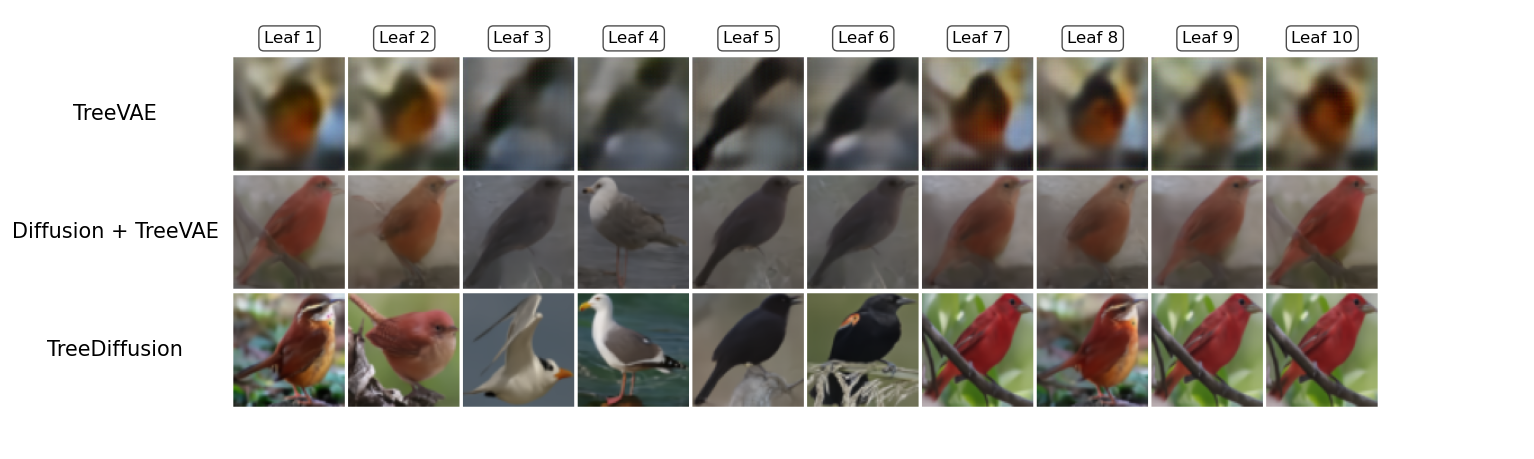}}
\centerline{\includegraphics[width=\textwidth, trim=0 20 90 0, clip]{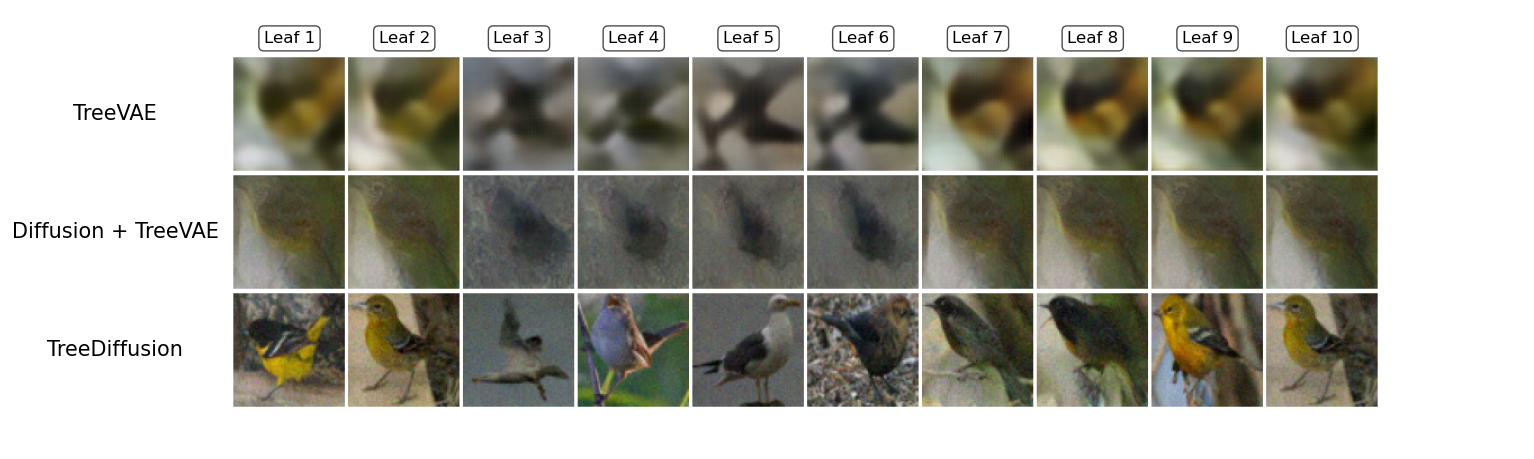}}
\caption{For each example, we show image generations from every leaf of TreeVAE, TreeVAE + Diffusion, and TreeDiffusion
model, all trained on CUBICC. Each row shows the generated images from all leaves
of the respective model, starting with the same root sample.}
\label{cubicc_generations_generations1}
\end{center}
\end{figure*}

\clearpage

\section{Implementation Details}

\subsection*{TreeVAE Training}

\begin{table}[htbp]
\centering
\caption[TreeVAE training configurations]{
Overview of training configurations for TreeVAE, including model parameters, training hyperparameters, and contrastive learning specifics across datasets.}
\label{tab:cnn_treevae_hyperparameters}
\begin{tabular}{lccc}
\toprule
\textbf{Data resolution} & 28x28x1 & 32x32x3 & 64x64x3 \\
\midrule
Encoder/Decoder Types & cnn1 & cnn1 & cnn2 \\
Max. tree depth & 7 & 7 & 7 \\
Max. clusters & 10 & 10 & 10 \\
Representation dimensions  & 4 & 4 & 4 \\
Latent channels & 16 & 64 & 64 \\
Bottom-up channels & 32 & 128 & 128 \\
Grow & True & True & True \\
Prune & True & True & True \\
Activation of last layer & sigmoid & mse & mse \\
\hline
Optimizer & Adam(lr=1e-3) & Adam(lr=1e-3) & Adam(lr=1e-3) \\
Effective batch size & 256 & 256 & 128 \\
Initial epochs & 150 & 150 & 150 \\
Smalltree epochs & 150 & 150 & 150 \\
Intermediate epochs & 80 & 0 & 0 \\
Fine-tuning epochs & 200 & 0 & 0 \\
lr decay rate & 0.1   & 0.1 & 0.1 \\
lr decay step size & 100 & 100 & 100 \\
Weight decay  & 1e-5  & 1e-5  & 1e-5  \\
\hline
Contrastive augmentations & False & True & True \\
Augmentation method & None & InfoNCE & InfoNCE \\
Augmentation weight & None & 100 & 100 \\
\bottomrule
\end{tabular}
\end{table}

We use a modified variant of the TreeVAE model, employing CNNs for its operations. The representations within TreeVAE are maintained in a 3D format, where the first dimension signifies the number of channels, while the subsequent two dimensions denote the spatial dimensions of the representations. Consequently, the deterministic variables in the bottom-up pathway possess a dimensionality of (Nb. of bottom-up channels, representation dimension, representation dimension), while the stochastic variables in the top-down tree structure have a dimensionality of (Nb. of latent channels, representation dimension, representation dimension). Table \ref{tab:cnn_treevae_hyperparameters} provides details on the remaining parameters utilized for the TreeVAE model and its training. For more information, please refer to the code.

\newpage

\subsection*{TreeDiffusion Training}

Table \ref{tab:ddpm_hyperparameters} provides details on the remaining parameters utilized for the diffusion model and its training. For more information, please refer to the code.

\begin{table}[ht]
\centering
\caption[TreeDiffusion training configurations]{
Overview of training configurations for the DDPM of the TreeDiffusion, including model parameters and training hyperparameters.}
\label{tab:ddpm_hyperparameters}
\begin{tabular}{lccc}
\toprule
\textbf{Data resolution} & 28x28x1 & 32x32x3 & 64x64x3 \\
\midrule
Noise Schedule & Linear(1e-4, 0.02) & Linear(1e-4, 0.02) & Linear(1e-4, 0.02) \\
U-Net channels & 64 & 128 & 128 \\
Scale(s) of attention block & [16] & [16,8] & [16,8] \\
Res. blocks per scale & 2 & 2 & 2 \\
Channel multipliers & (1,2,2,2) & (1,2,2,2) & (1,2,2,2,4) \\
Dropout & 0.3 & 0.3 & 0.3 \\
\hline
Diffusion loss type & L2 & L2 & L2\\
Optimizer & Adam(lr=2e-4) & Adam(lr=2e-4) & Adam(lr=2e-4) \\
Effective batch size & 256 & 256 & 32 \\
lr annealing steps & 5000 & 5000 & 5000 \\
Grad. clip threshold & 1.0 & 1.0 & 1.0 \\
EMA decay rate & 0.9999 & 0.9999 & 0.9999 \\
\bottomrule
\end{tabular}
\end{table}

\end{document}